\documentclass[11pt]{article}

\usepackage[preprint]{acl}

\usepackage{times}
\usepackage{latexsym}

\usepackage[T1]{fontenc}
\usepackage[utf8]{inputenc}

\usepackage{microtype}

\usepackage{inconsolata}

\usepackage{graphicx}
\usepackage{xspace}
\usepackage{booktabs} 
\usepackage{multirow} 
\usepackage{url}
\usepackage{xurl}
\usepackage{amssymb}
\usepackage{tabularx}
\usepackage[most]{tcolorbox}

\newcommand{\system}{{\textsc{SpeechEQ}}\xspace}

\newtcolorbox{promptbox}[1][]{
  colback=gray!5!white,
  colframe=gray!75!black,
  fonttitle=\bfseries,
  coltitle=black,
  colbacktitle=gray!20!white,
  enhanced,
  breakable, % Allows the box to split across pages
  title=#1
}

\title{\system: Benchmarking  Emotional Intelligence Quotient in Socially Aware Voice Conversational Models}

\author{
Liang-Yuan Wu\textsuperscript{1},
Zih-Ching Chen\textsuperscript{2},
Tongshuang Wu\textsuperscript{3},
C.-H. Huck Yang\textsuperscript{2},
Hua Shen\textsuperscript{1, 4} \\
\textsuperscript{1}New York University \quad\textsuperscript{2}NVIDIA \quad \textsuperscript{3}Carnegie Mellon University \quad
\textsuperscript{4}NYU Shanghai\\
\texttt{\{leo.wu,hua.shen\}@nyu.edu}; \texttt{\{virginiac,hucky\}@nvidia.com}; \texttt{sherryw@cs.cmu.edu} 
}

\begin{document}
\maketitle
\begin{abstract}
Paralinguistic cues that convey social intent are critical for spoken conversational systems. However, existing evaluations typically focus on text or isolated emotion recognition, leaving the social meaning and contextual appropriateness of vocal delivery underexamined. We introduce \system, a controlled diagnostic benchmark comprising 2,265 multi-turn dialogues organized by the 15 EQ-i 2.0 Emotional Intelligence Assessment. At each of two decision turns, a model selects between spoken responses with identical transcripts but different vocal deliveries. We report trajectory accuracy as the absolute measure of joint success and the cohort-normalized Spoken Emotional Quotient (SEQ) index as a centered view of relative performance. Across cascaded and end-to-end Speech-Language Models (SLMs), performance varies substantially: the strongest end-to-end system outperforms the cascaded baselines, while several native-audio models remain near chance. Analysis of the best-performing model, Qwen3-Omni-30B, reveals arousal aversion, multi-turn performance degradation, and asymmetric sensitivity to EQ-specific persona conditioning. A human-ranking study further shows strong agreement between benchmark scores and human judgments of emotional intelligence, supporting the validity of our evaluation metric. Demos and data are available on our project website\footnote{\url{https://speecheq.github.io}}.
\end{abstract}

\section{Introduction}
\begin{figure}[!t]
\begin{center}
%\framebox[4.0in]{$\;$}
% \fbox{\rule[-.5cm]{0cm}{4cm} \rule[-.5cm]{12cm}{0cm}}
\includegraphics[width=\linewidth]
{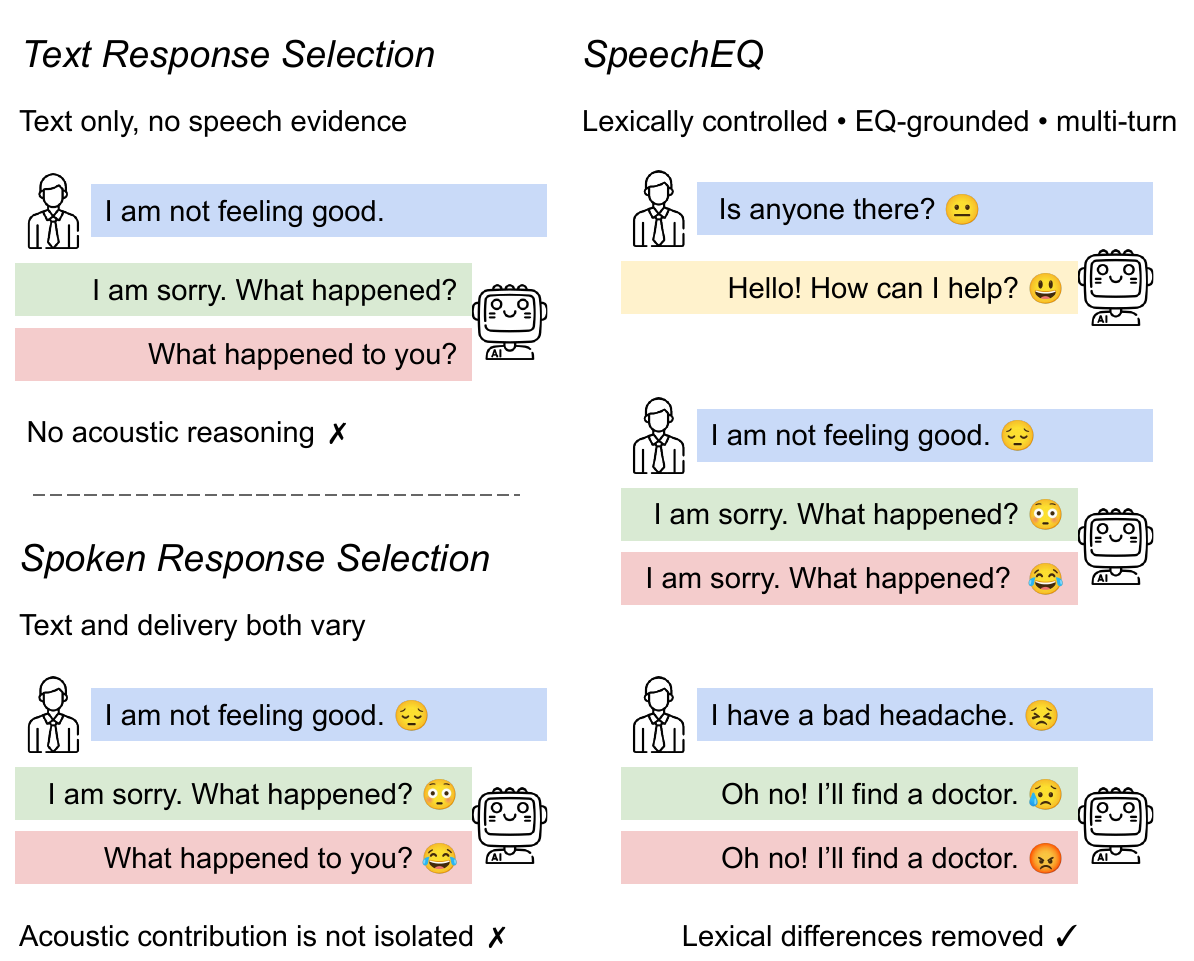}
\end{center}
\vspace{-10pt}
\caption{Overview of \system's controlled paralinguistic response-selection task for SLMs.}
\label{fig:overview}
\vspace{-10pt}
\end{figure}

Recent advances in Speech-Language Models (SLMs) have enabled fluent~\cite{defossez2024moshi}, dynamic interaction~\cite{rubenstein2023audiopalm, zhang2023speechgpt, chu2024qwen2} with voice agents. However, \textbf{effective spoken communication is not purely semantic}: paralinguistic cues such as intonation, pacing, and voice quality convey social and affective information~\cite{scherer2003vocal, wu2025soundnarratives, kim2023visible}. Despite their strong semantic capabilities, current SLMs struggle to interpret these cues~\cite{qian2025prosodylm}. This limits their effectiveness when understanding an interaction requires recognizing empathy, tension, or intent conveyed through vocal delivery, highlighting the need to evaluate Emotional Intelligence (EQ)~\cite{salovey1990emotional, elfenbein2002universality} in spoken communication.

We argue that this limitation stems from a deeper issue: \textbf{the lack of rigorous evaluation for Emotional Intelligence in speech}. Existing benchmarks either (i) evaluate emotional intelligence in text-only settings \cite{paech2023eq, sabour2024emobench} or (ii) treat speech as a passive perception task \cite{schuller2018speech, yang2021superb}. Recent benchmarks extend evaluation to emotionally rich, multi-turn spoken dialogue~\cite{deng2025multi,wang2026humdial,koudounas2025deepdialogue}. However, when candidate responses differ in both wording and vocal delivery, the contribution of acoustic reasoning is difficult to isolate. Models may achieve high performance while relying on a ``\textbf{semantic shortcut}'', bypassing acoustic reasoning altogether.

\begin{figure*}[!t]
\begin{center}
%\framebox[4.0in]{$\;$}
% \fbox{\rule[-.5cm]{0cm}{4cm} \rule[-.5cm]{12cm}{0cm}}
\includegraphics[width=\linewidth]
{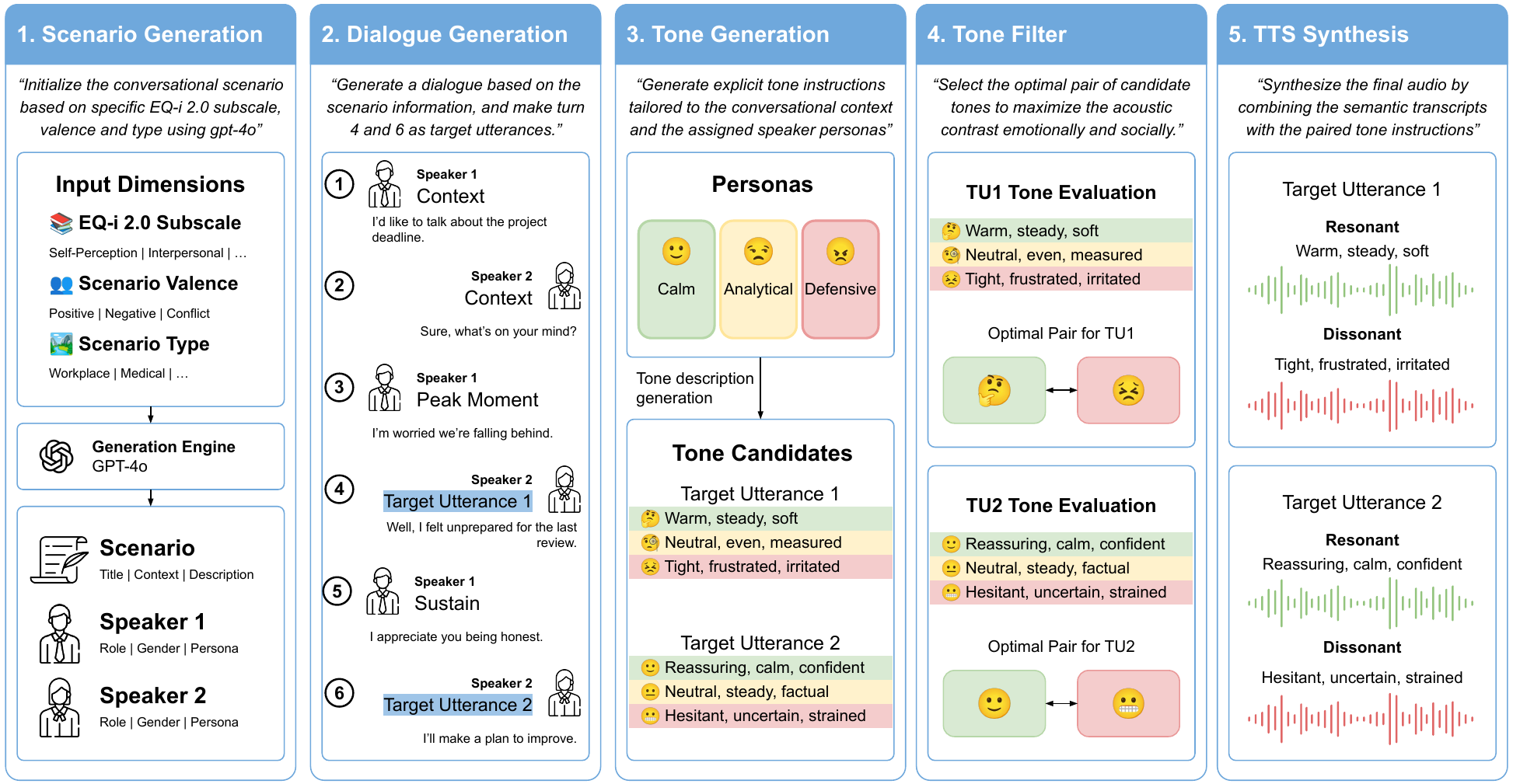}
\end{center}
\vspace{-10pt}
\caption{Overview of \system construction pipeline.}
\label{fig:pipeline}
\vspace{-10pt}
\end{figure*}

To address this gap, we introduce \system (Figure \ref{fig:overview}), a benchmark for \textbf{context-conditioned paralinguistic reasoning in spoken dialogue}. \system comprises 2,265 six-turn dialogues (42.37 hours) grounded in the 15 EQ-i 2.0 subscales~\cite{bar2004bar,wiechorek2011emotional}. It is built on three key principles: 
\textbf{(i) Behavioral Grounding via EQ-i 2.0.}
We construct scenarios that map psychological constructs (e.g., empathy, impulse control) into observable acoustic behaviors.
\textbf{(ii) Semantic–Acoustic Decoupling.}
We isolate acoustic reasoning by designing response options that share identical transcripts but differ in paralinguistic delivery. This removes lexical differences between candidates, requiring judgements of vocal delivery in context.
\textbf{(iii) Sustained Affective Pragmatics.}
We evaluate multi-turn dialogues with escalating emotional stakes, testing whether models can track and adapt to evolving social dynamics over time.
The data are generated via a controlled pipeline (Figure \ref{fig:pipeline}) that enforces both behavioral validity and acoustic contrast. To quantify performance, we introduce the Spoken Emotional Quotient (SEQ), a cohort-relative index derived from trajectory accuracy. It measures immediate recognition and sustained emotional reasoning. A human-ranking study shows it correlates strongly with human preferences, and provides evidence of response-selection rationales.

Using \system, we benchmark both cascaded pipelines and state-of-the-art end-to-end SLMs, observing substantial variation in performance, latency, and cost. The strongest end-to-end model outperforms the cascaded baselines, while several native-audio models remain near chance. Further analysis of the strongest-performing model reveals three challenges:
(i) \textbf{Arousal Aversion.} Performance declines when socially appropriate responses require greater arousal.
(ii) \textbf{Multi-Turn Degradation.} Performance is lower in sequential interaction than when the same dialogue history is evaluated in a single inference.
(iii) \textbf{Asymmetric Persona Effects.} EQ-specific deficit personas affect performance unevenly, with stress-management deficits causing the largest decline.
Together, these findings highlight the need to evaluate context-sensitive vocal judgments across emotional demands and interaction settings.

Our contributions are three-fold: (1) \system, an EQ-grounded benchmark containing 2,265 multi-turn spoken dialogues with lexically controlled response pairs; (2) a two-stage evaluation protocol with trajectory accuracy and a cohort-normalized SEQ index for comparing sustained paralinguistic appropriateness; and (3) a broad comparison of cascaded and end-to-end SLMs that reveals substantial performance variation and turn-dependent degradation.

\section{\system Benchmark}

\subsection{Motivation and Design Rationale}

\textbf{Attributable Behavior Design.} Our goal is to move beyond emotion classification in traditional speech emotion recognition \cite{cowie2001emotion, burkhardt2000database, schuller2018speech} toward evaluating context-dependent social judgments in spoken dialogue. We design response pairs whose appropriateness can be distinguished through vocal delivery while holding their lexical content constant. We ground our dataset in the EQ-i 2.0 framework \cite{bar2004bar, wiechorek2011emotional}. EQ-i 2.0 is a trait-behavioral model that operationalizes social functioning into measurable subscales. This framework guides the construction of scenarios linking social and emotional behaviors to contextually appropriate vocal delivery.

\textbf{Tone Variation Design.} To mitigate lexical bias \cite{chen2026audio, wang2020makes} and rigorously evaluate acoustic emotional intelligence, we designed the task as a forced-choice selection between two audio responses sharing identical transcripts. By neutralizing the text modality, we eliminate semantic differences and force the system to evaluate subtle paralinguistic cues to determine the contextually resonant response.

\textbf{Multi-turn Conversational Arc.} To capture emotional intelligence beyond single utterances, we structure each scenario as three exchanges between a user and a model. The first exchange establishes the emotional context. In the second and third exchanges, the model selects between two candidate spoken responses as the user's emotional situation develops. This design evaluates context-dependent judgments of vocal delivery across an evolving dialogue.

\subsection{Data Generation}
\label{sec:data_generation}
We developed an automated, LLM-driven pipeline for scenario and tone generation. Figure \ref{fig:pipeline} presents the complete five-stage pipeline; here, we highlight how we construct EQ-grounded scenarios and create contrastive tone pairs.

\textbf{Scenario Generation and Persona Matrix.} We construct a constrained scenario matrix under the EQ-i 2.0 taxonomy. Each scenario lies at the intersection of an EQ-i 2.0 subscale (e.g., interpersonal relationships or empathy), a situational valence (positive, negative, or conflict), and a setting (e.g., workplace, medical, or educational). The pipeline also generates contrasting social-behavior personas for the targeted subscale, such as a ``toxic optimist'' who fails to validate grief. These personas guide the creation of contextually appropriate and inappropriate vocal deliveries.

\textbf{Tone Generation and Contrast Filtering.} We prompt an LLM to translate three personas (one contextually appropriate response and two inappropriate distractors) into physically grounded vocal instructions for the same transcript. To avoid low-contrast descriptions, a filter rejects generic descriptors such as ``polite'' or ``calm.'' An automated judge then retains the contextually appropriate instruction and the most contrasting distractor for synthesis with \verb+gpt-4o-mini-tts-2025-03-20+. The resulting candidate pair therefore differs in delivery while holding lexical content constant.

\subsection{Data Validation}

% To ensure the quality of \system, we adopt a two-phase validation pipeline.
% Phase 1 (automated) verifies scenario consistency and acoustic distinctiveness, while Phase 2 (human) evaluates naturalness and perceptual validity.

% \begin{table}[h]
% \small
% \begin{center}
% \begin{tabular}{llll}
% \toprule
%  & $Acc_1$ & $Acc_2$ & $Acc_{traj}$ \\
% \midrule
% Qwen2.5-3B & 98.6\%    & 99.1\%   & 97.7\% \\
% Qwen2.5-7B & 99.6\%    & 99.9\%   & 99.5\% \\
% Qwen3-30B & 99.7\%    & 100.00\%   & 99.7\% \\
% \bottomrule
% \end{tabular}
% \end{center}
% \vspace{-8pt}
% \caption{Semantic validation results.}\label{tab:semantic_eval}
% \vspace{-10pt}
% \end{table}

\textbf{Semantic and Logical Verification.}
We first conduct an \emph{oracle} text-based evaluation by providing models with transcripts and explicit tone instructions instead of audio (e.g., \textit{``Fast pacing, sarcastic cheerfulness.''}). Models achieve 97.7\% accuracy on Qwen2.5-3B and over 99.5\% on Qwen2.5-7B and Qwen3-30B. These results suggest that the audio task’s difficulty is not primarily explained by ambiguity in the intended choices, and we isolate the challenge of \system from understanding audio paralinguistics cues.

% We first conducted an ``oracle'' text-based evaluation by providing the Qwen model family with raw conversational transcripts and explicit TTS tone instructions (e.g., \textit{``Fast pacing, sarcastic cheerfulness.''}) instead of audio. In this text-only modality, the models achieved near-perfect accuracy (Table \ref{tab:semantic_eval}). This confirms that the generated scenarios are logically consistent and unambiguously align with the 15 emotional intelligence subscales of the EQ-i 2.0 framework \cite{bar2004bar, wiechorek2011emotional} (see Appendix \ref{app::eqi_2} for the complete taxonomy). Consequently, this proves that the difficulty of \textsc{SpeechEQ} stems entirely from the modality gap—the challenge of extracting paralinguistics from audio—rather than from ambiguous scenario generation.

\textbf{Acoustic Variance Validation.} 
To ensure meaningful paralinguistic contrast, we quantify acoustic differences between resonant and dissonant clips for each evaluation pair (Turns 4 and 6) using librosa. We extract six dimensions: mean pitch, zero-crossing rate (speaking-rate proxy), spectral centroid, RMS energy, mean MFCC, and duration. We compute a composite contrast score (max 8 points), with pitch and speaking-rate gaps contributing up to 2 points each and the remaining four up to 1 point each. Pairs scoring below 4 trigger up to three TTS regeneration attempts; examples that still fail are discarded.

% To verify that our synthesis engine successfully generated distinct paralinguistic variations, we mathematically quantified the acoustic differences between the resonant and dissonant audio options. For each semantically identical evaluation pair (Turns 4 and 6), we extracted a suite of low-level acoustic features, including pitch, speech rate, spectral shape, energy, and absolute duration. We then calculated the relative variance between the feature vectors of the high-EQ and low-EQ audio tracks and established a minimum acoustic distinction threshold. Only evaluation pairs that surpassed this threshold were preserved in the final dataset. This strict filtering guarantees that every scenario in \textsc{SpeechEQ} presents a verifiable acoustic mismatch, ensuring the necessary variance required to rigorously evaluate end-to-end language models.

\textbf{Human Expert Validation.}
We assess perceptual validity through expert annotation of 75 randomly sampled scenarios (5 per EQ subscale). Two expert annotators independently evaluate five criteria covering text and audio quality, semantic and acoustic EQ relevance, and answer correctness. Table~\ref{tab:valid_res} reports results before filtering; metric definitions and literature grounding appear in Table~\ref{tab:metric_def} in Appendix~\ref{app::human_metric}. Inter-annotator agreement before reconciliation is substantial (Cohen's $\kappa = 0.617$), and annotation results support the quality and perceptual validity of the reviewed sample. Disagreements are subsequently resolved through discussion, and scenarios within this sample that fail any criterion are removed from the dataset.

% While automated metrics confirm mathematical variance and logical solvability, human evaluation is required to verify social naturalness and perceptual validity. We sampled a subset of 5 scenarios per EQ subscale, yielding 75 total evaluation instances. Two AI domain experts annotated these instances over two rounds across five key dimensions: Generation Quality (Text and Audio), EQ Relevance (Text and Audio), and Answer Correctness (Paralinguistic Accuracy). Formal definitions and literature grounding for these metrics are provided in Table \ref{tab:metric_def} in Appendix \ref{app::human_metric}. The evaluation results are summarized in Table \ref{tab:valid_res}. Disagreements between annotators were resolved through iterative discussion, achieving strong inter-rater reliability (Cohen's Kappa $\kappa = 0.617$). Any scenario failing a single evaluation dimension was categorically filtered from the final benchmark.

\textbf{Data Statistics.} The dataset comprises a total of 2,265 dialogues, balanced across the 15 EQ-i 2.0 subscales, totaling 42.37 hours of audio. The mean dialogue duration is 67.35 seconds ($\sigma=22.29$). 

% \begin{table*}[h]
% \footnotesize
% \centering
% % \begin{tabular}{@{}l|ll|ll|l@{}}
% \begin{tabular}{p{0.12\linewidth}|p{0.1\linewidth}p{0.12\linewidth}|p{0.04\linewidth}p{0.14\linewidth}|p{0.13\linewidth}}
  
% \toprule
% \textbf{Categories} & \multicolumn{2}{l|}{\textbf{Generation Quality}}           & \multicolumn{2}{l|}{\textbf{EQ Relevance}}                                & \textbf{Answer Correctness}      \\ \midrule
% Metrics    & \multicolumn{1}{l|}{Text Quality}                  & Audio Quality                 & \multicolumn{1}{l|}{Semantic Relevance}       & Acoustic Reasonability       & Paralinguistic Accuracy                                       \\ \midrule
% Results    & \multicolumn{1}{l|}{1.00}                          & 0.98                          & \multicolumn{1}{l|}{0.93}                     & 0.98                         & 0.94                                                          \\ \bottomrule
% \end{tabular}
% \caption{Data validation results from human experts.}\label{tab:valid_res}
% \vspace{-12pt}
% \end{table*}

\begin{table}[t]
\small
\centering
\begin{tabular}{@{}llc@{}}
\toprule
\textbf{Category} & \textbf{Criterion} & \textbf{Score} \\
\midrule
\multirow{2}{*}{Generation quality}
    & Text quality  & 1.00 \\
    & Audio quality & 0.98 \\
\addlinespace[2pt]
\multirow{2}{*}{EQ relevance}
    & Semantic relevance      & 0.93 \\
    & Acoustic reasonability  & 0.98 \\
\addlinespace[2pt]
Answer correctness
    & Paralinguistic accuracy & 0.94 \\
\bottomrule
\end{tabular}
\caption{Expert validation scores across five criteria.}
\label{tab:valid_res}
\vspace{-8pt}
\end{table}

\section{Evaluation Framework}
\subsection{Evaluation Protocol for EQ}

\textbf{Two-Round Selection Process.} We evaluate models through a sequential forced-choice task at Turns 4 and 6 of each dialogue. At each decision turn, the model selects between two audio candidates with identical transcripts but contrasting vocal deliveries. In Round 1, a user prompt provides the scenario context, Turns 1--3 audio, and the Turn-4 candidates. The model returns its selection and rationale. In Round 2, we append a new user prompt to this conversation, retaining the first prompt and the model's reply. The new prompt includes the selected Turn-4 audio, the subsequent Turn-5 utterance, and the Turn-6 candidates. The model then returns its second selection. The selected Turn-4 response is used regardless of correctness, allowing us to evaluate context-sensitive judgments across successive decisions. Full evaluation prompts are provided in Appendix~\ref{app::evaluation_details}.

\textbf{Evaluation Metrics.} $Acc_1$ and $Acc_2$ measure selection accuracy at Turns 4 and 6 respectively. We report conversational trajectory accuracy ($Acc_{traj}$) to measure sustained performance across the dialogue. This requires the model to select the socially appropriate response at both evaluation turns. For a dataset of $N$ multi-turn scenarios, let $\hat{y}_{i,1}$ and $\hat{y}_{i,2}$ denote the model's predicted choices for the $i$-th conversation at Turns 4 and 6, with $y_{i,1}$ and $y_{i,2}$ representing the respective ground-truth resonant labels. The sustained accuracy is formally defined as the joint success across both evaluation turns, utilizing the indicator function $\mathbb{I}$:

\vspace{-12pt}
\begin{equation}
    \label{equ::acc}
    Acc_{traj} = \frac{1}{N} \sum_{i=1}^{N} \mathbb{I}(\hat{y}_{i,1} = y_{i,1} \land \hat{y}_{i,2} = y_{i,2})
\end{equation}
\vspace{-10pt}

Uniform random selection yields expected values of 50\%, 50\%, and 25\% for $Acc_1$, $Acc_2$, and $Acc_{traj}$. We use $Acc_{traj}$ for absolute cross-model comparison.

\subsection{Spoken Emotional Quotient (SEQ)}

$Acc_{traj}$ provides the absolute performance measure. To make relative differences within the evaluated model cohort easier to inspect, we additionally report the cohort-normalized Spoken Emotional Quotient (SEQ) index.

\textbf{Global Standardization.} For model $i$, let $X_i=Acc_{traj}$ denotes its trajectory accuracy. We standardize $X_i$ relative to the cohort using the Median Absolute Deviation ($\mathrm{MAD}$) as a robust statistical measure for a resilient standardization~\cite{leys2013detecting}. 

\vspace{-10pt}
\begin{equation}
    \label{equ::z_score}
    Z_{i} = \frac{X_i - \text{Median}(X)}{k \times \text{MAD}(X)}
\end{equation}
\vspace{-8pt}

where $k \approx 1.4826$ is the standard scaling factor. This constant is derived from the inverse of the 75th percentile of the standard normal distribution ($1 / \Phi^{-1}(0.75)$), which ensures the $\mathrm{MAD}$ is asymptotically consistent with the standard deviation of a normal distribution \citep{rousseeuw1993alternatives}.

\textbf{Final SEQ Computation.} Following standard clinical psychometric scaling, we map the resulting values to a distribution centered at $\mu$ with scale $\sigma$ and cap scores at $\pm 4 \sigma$~\cite{wechsler1955wechsler},  reflecting the
floor and ceiling limits of standardized assessments~\cite{anastasi1988psychological}. The final SEQ score is:

\vspace{-10pt}
\begin{equation}
    \label{equ::seq_raw}
    \text{SEQ}_i = \operatorname{clip}_{[\mu-4\sigma,\mu+4\sigma]}(\mu + \sigma \times Z_i)
\end{equation}
\vspace{-10pt}

% \vspace{-6pt}
% \begin{equation}
% \label{equ::seq}
% \mathrm{SEQ}_i =
% \operatorname{clip}_{[\mu-4\sigma,\mu+4\sigma]}
% \left(
% \mu + \sigma \cdot
% (\frac{X_i-\operatorname{median}(X)}
% {k \times\,\mathrm{MAD}(X)})
% \right)
% \end{equation}
% \vspace{-6pt}

We set $\mu = 100$ and $\sigma = 15$ to follow standardized cognitive and emotional intelligence frameworks \cite{bar2004bar, wiechorek2011emotional}.

\subsection{Experimental Settings}

We evaluate cascaded pipelines and end-to-end SLMs under the same response-selection protocol to establish rigorous baselines for \system.

\textbf{Cascaded Systems:} To establish a lower bound simulating systems, we use two models to extract lexical content and affective information from vocal delivery. \path{Whisper-large-v3} ASR~\cite{radford2023robust} transcribes each candidate audio, while a pretrained SER model\footnote{\path{audeering/wav2vec2-large-robust-12-ft-emotion-msp-dim}}~\cite{wagner2023dawn} estimates its valence, arousal, and dominance (VAD).  We augment the transcripts with either raw numerical VAD estimates or categorical tone descriptions derived from them. The text-only LLM (\verb+Qwen3-30B+~\cite{yang2025qwen3}) receives the scenario background and augmented candidate transcripts, then selects the response whose described vocal delivery is more appropriate for the context.

\textbf{End-to-End SLMs:} We evaluate open-weight models across a range of scales: the Qwen-Omni series \cite{xu2025qwen3}, Kimi-Audio-7B-Instruct \cite{ding2025kimi}, MiMo-Audio-7B-Instruct  \cite{zhang2025mimo}, and Fun-Audio-Chat-8B \cite{team2025fun}. We also evaluate two commercial APIs: Gemini-2.5-Pro \cite{comanici2025gemini} and gpt-audio-1.5 \cite{openai2026gptaudio15}. For all end-to-end models, the scenario background is provided as text, while the dialogue history and candidate response options are interleaved into the context window as native audio clips.

\section{Results}
\begin{table*}[h]  
\small
\centering
\begin{tabular}{@{}l c c c c c c c@{}}
\toprule
\multirow{2}{*}{\textbf{Model}} & \multicolumn{4}{c}{\textbf{Selection Performance}} & \multicolumn{3}{c}{\textbf{Observed Deployment Characteristics}} \\ \cmidrule(lr){2-5} \cmidrule(l){6-8}
 & $\boldsymbol{Acc_1}$ & $\boldsymbol{Acc_2}$ & $\boldsymbol{Acc_{traj}}$ & \textbf{SEQ} & \textbf{Cost (\$/100 ex.)} & \textbf{Latency (s)} & \textbf{Speed (tok/s)} \\ \midrule
Uniform Random & 0.500 & 0.500 & 0.250 & 88.39 & -- & -- & -- \\ \midrule
\multicolumn{8}{@{}l}{\textit{Cascaded Pipelines / Voice Agents}} \\ \midrule
$emo_{num}$ + Qwen3-30B & 0.569 & 0.536 & 0.358 & 107.22 & 0.47 & 20.9 & 189.2 \\
$emo_{des}$ + Qwen3-30B & 0.606 & 0.593 & 0.403 & 115.29 & 0.47 & 21.1 & 189.8 \\ \midrule
\multicolumn{8}{@{}l}{\textit{End-to-End SLMs}} \\ \midrule
Qwen2.5-Omni-3B & 0.556 & 0.548 & 0.306 & 98.12 & 0.20 & 8.8 & 615.7 \\
Qwen2.5-Omni-7B & 0.508 & 0.502 & 0.260 & 89.88 & 0.17 & 7.9 & 688.0 \\
Qwen3-Omni-30B & \textbf{0.785} &\textbf{0.708} & \textbf{0.583} & \textbf{147.26} & 0.46 & 20.8 & 187.4 \\
Kimi-Audio-7B-Instruct & 0.501  & 0.481  & 0.242     & 86.59  & 0.31 & 13.8    & 202.4 \\
MiMo-Audio-7B-Instruct & 0.509  & 0.519  & 0.271     & 91.76  & 0.23 & 10.4    & 200.6 \\
Fun-Audio-Chat-8B      & 0.681  & 0.528  & 0.365     & 108.55 & 0.22 & 9.8     & 270.4 \\
Gemini-2.5-pro         & 0.683  & 0.639  & 0.449     & 123.45 & 1.12 & 29.3    & 129.1 \\
gpt-audio-1.5 & 0.555 & 0.545 & 0.317 & 100.00 & 5.63 & 8.7 & 392.2 \\ \bottomrule
\end{tabular}
\caption{\system evaluation results. SEQ is normalized within the reference cohort. Deployment values are observations under the authors' setup.}
\label{tab:baseline}
\vspace{-10pt}
\end{table*}

\subsection{How do cascaded and end-to-end systems compare?}

\textbf{State-of-the-art SLM outperforms its cascaded counterparts (Table \ref{tab:baseline}), demonstrating the broader limitations of traditional SER models.} The end-to-end \verb+Qwen3-Omni-30B+ model (\textit{i.e.}, processes continuous speech directly) achieves a substantially higher accuracy. Within the cascaded systems, mapping continuous VAD estimates to descriptive tone labels ($emo_{des}$) improves over appending raw values ($emo_{num}$). However, native-audio processing does not confer a uniform advantage: several end-to-end systems remain close to the 0.25 uniform-random trajectory baseline. The results therefore show substantial variation by model rather than architecture-wide ordering.

\textbf{Within end-to-end SLMs, we observe trade-offs in performance, latency, and operational cost (Table \ref{tab:baseline}).} We quantified these metrics via unbatched, single-stream inference with 100-example runs on an NVIDIA A100 GPU for open-weight models, compared against online APIs. While the 30B Qwen3-Omni model dominates in raw EQ performance, it suffers from high latency, taking 2.5$\times$ longer to respond than its smaller counterparts. Within the smaller models that offer fast, highly cost-effective inference, Fun-Audio-Chat-8B has the best performance.

% \begin{figure*}[h]
% \begin{center}
% \includegraphics[width=\linewidth]
% {figures/SEQ_chart_group_1.png}
% \end{center}
% \vspace{-15pt}
% \caption{SEQ score for different cascaded systems and E2E SLMs.}
% \label{fig:SEQ}
% \vspace{-12pt}
% \end{figure*}

\subsection{Do models favor calmer delivery when higher arousal is appropriate?}

\begin{figure}[!t]
\begin{center}
\includegraphics[width=\linewidth]
{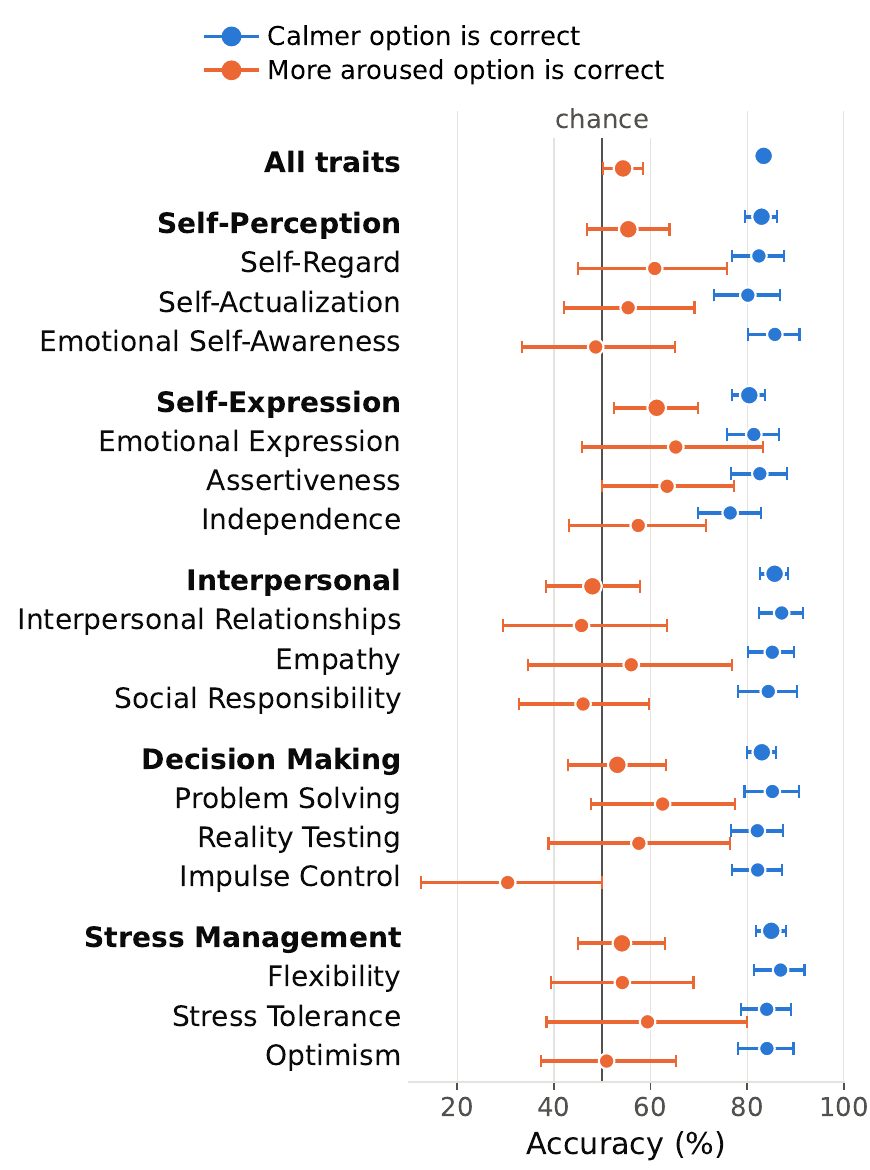}
\end{center}
\vspace{-15pt}
\caption{Qwen3-Omni-30B accuracy when the low versus high arousal candidate is contextually appropriate.}
\label{fig:arousal}
\vspace{-12pt}
\end{figure}

\textbf{Qwen3-Omni-30B is less accurate when socially appropriate delivery requires greater arousal (Figure \ref{fig:arousal}).} Emotionally appropriate responses do not always equal to calm, low arousal signals, and some EQ scenarios call for greater arousal. We examine whether Qwen3-Omni-30B, the model with the best performance, favors low arousal candidates even when the more aroused response is appropriate in the scenario.

We estimate relative arousal from audio loudness, pitch variability, and speaking rate, excluding pairs with small differences. This yields 2,758 pairs where the calmer candidate is more appropriate and 604 where the more aroused candidate is more appropriate.

Accuracy is 83.4\% when the candidate with lower arousal is correct, but drops to 54.3\% when the more aroused candidate is correct. Figure~\ref{fig:arousal} shows this asymmetry across EQ composite areas and subscales. Notably, accuracy remains near chance for Optimism (50.9\%) when the appropriate response is more aroused. In such contexts, conveying optimism may require energetic encouragement rather than restrained delivery, yet the model struggles to select that response.

\subsection{How does the inference setting affect multi-turn performance?}

\textbf{The Turn-6 performance drop depends on the interaction setup.} Qwen3-Omni-30B achieves 78.5\% accuracy at Turn 4 but only 70.8\% at Turn 6 under the standard two-round interaction. We first investigate whether Turn-6 items are inherently more difficult by evaluating them in a fresh conversation using the Turn-4 instructions, with S3--S5 as context and the correct S4 response. Performance largely recovers under this local-window condition, suggesting that item difficulty alone does not explain the observed gap.

We next examine whether errors in the preceding response account for the degradation. Retaining the multi-turn interaction while replacing the selected S4 response with the correct one yields only a small improvement. Finally, we consolidate the full dialogue history (S1--S5, with correct S4) into a single user message, without an intermediate Turn-4 assistant reply or rationale. Accuracy rises from 72.1\% in the corrected-S4 multi-turn condition to 75.0\% in the single-message condition. All Turn-6 conditions use the same 2,265 items, with candidate order, scenario descriptions, and sampling settings held constant. Together, these results suggest that the multi-turn interaction setup contributes to the degradation beyond item difficulty or preceding-response errors, motivating further investigation into reliable emotional reasoning across successive interactions.

\begin{figure}[h]
\begin{center}
\includegraphics[width=\linewidth]
{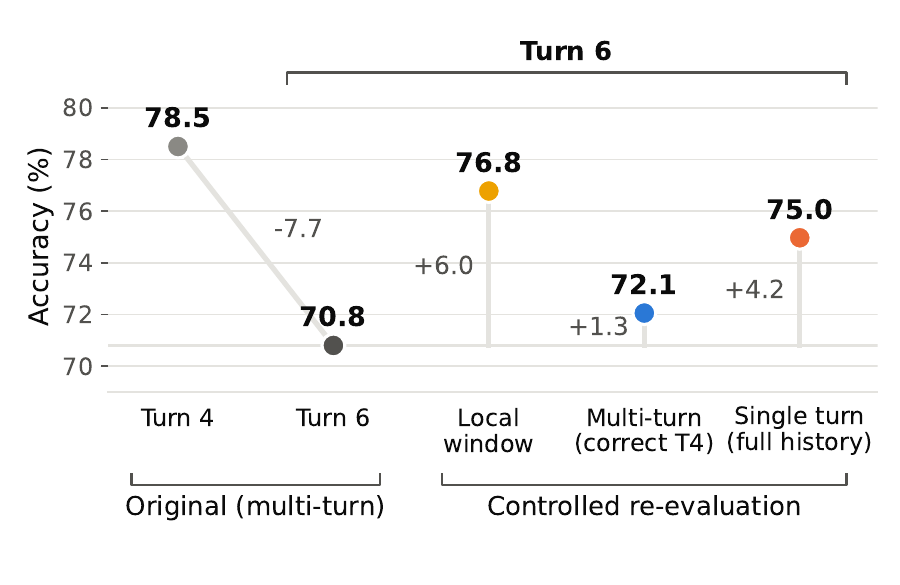}
\end{center}
\vspace{-12pt}
\caption{Qwen3-Omni-30B accuracy under the original evaluation and controlled Turn-6 conditions.}
\label{fig:history}
\vspace{-12pt}
\end{figure}

\subsection{How does persona conditioning affect performance across EQ dimensions?}

\begin{figure*}[h]
\begin{center}
\includegraphics[width=\linewidth]
{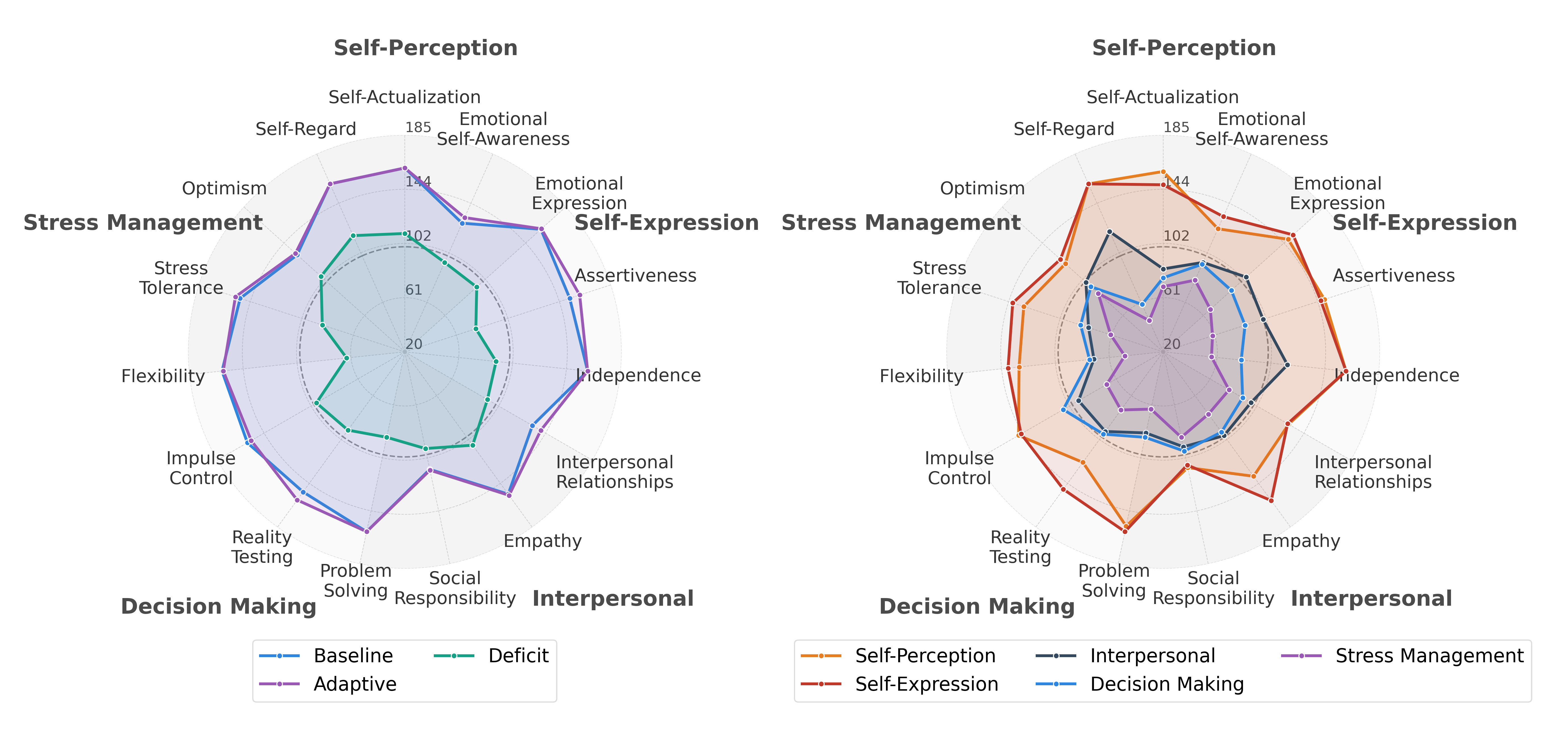}
\end{center}
\vspace{-15pt}
\caption{SEQ under different persona conditions (Qwen3-Omni-30B).}
\label{fig:SEQ_persona}
\vspace{-12pt}
\end{figure*}

\textbf{Persona conditioning affects response selection unevenly across EQ dimensions.} For Qwen3-Omni-30B, an emotionally adaptive persona yields a modest increase over the default prompt (SEQ: 147.26 to 148.86), whereas a global deficit persona substantially reduces performance (94.98; Figure~\ref{fig:SEQ_persona}, left). These results show that persona instructions can influence judgments of socially appropriate vocal delivery.

Targeted deficit personas produce markedly different effects. Self-Perception and Self-Expression deficits yield smaller declines (SEQ: 133.72 and 140.16), while the Stress Management deficit produces the largest decline (74.90; Figure~\ref{fig:SEQ_persona}, right). One possible explanation is that these personas prescribe different vocal preferences: reduced self-expression favors restrained, accommodating delivery, whereas poor stress management encourages reactive, aggressive responses under tension. Their effects may therefore depend on how these preferences match the social demands of the benchmark. Because the prompts differ in wording and behavioral prescriptions, the observed asymmetry does not isolate sensitivity to individual EQ traits or establish an alignment-related cause. Full persona prompts are provided in Appendix~\ref{app::persona_prompts}.

\subsection{Does \system align with human perception?}

We conduct a human-ranking study using one dialogue from each of the 15 EQ subscales. Six models select a candidate response and provide a written rationale. 15 native speakers recruited through Prolific~\cite{palan2018prolific} rank the anonymized rationales by perceived emotional intelligence. We compare SEQ with per-turn accuracy and pooled accuracy, which treats all Turn-4 and Turn-6 decisions as individual items. In contrast, SEQ derives from trajectory accuracy, requiring both decisions in a dialogue to be correct. Table~\ref{tab:human_ranking} shows stronger agreement with human rankings for SEQ ($\rho=0.943$) than for pooled accuracy ($\rho=0.886$), supporting joint success across turns as an indicator of human judgments of response-selection reasoning.

\begin{table}[h]
\begin{center}
\footnotesize
\begin{tabular}{@{}llllll@{}}
\toprule
            & Human & $Acc_1$ & $Acc_2$ & $Acc_{\mathrm{all}}$   & SEQ   \\ \midrule
$Model_A$    & 1     & 1       & 1=     & 1     & 1     \\
$Model_B$    & 3     & 2       & 3=     & 2     & 2     \\
$Model_C$    & 6     & 6       & 5=     & 6     & 6     \\
$Model_D$    & 2     & 3=      & 1=     & 3     & 3     \\
$Model_E$    & 4     & 3=      & 5=     & 5     & 4     \\
$Model_F$    & 5     & 3=      & 3=     & 4     & 5     \\ \midrule
correlation $\rho$ ($\uparrow$) & -     & 0.820   & 0.837  & 0.886 & \textbf{0.943} \\
$p$-value ($\downarrow$)     & -     & 0.046   & 0.039  & 0.018 & \textbf{0.005} \\ \bottomrule
\end{tabular}
\end{center}
\caption{Human and metric-based rankings of six anonymized models.}\label{tab:human_ranking}
\vspace{-12pt}
\end{table}

\section{Discussion}
\system highlights three challenges for socially aware speech models: interpreting acoustic cues, adapting emotional intensity to context, and sustaining emotional judgments across turns.

\textbf{Overcoming the Modality Shortcut}. Existing models can prioritize text over acoustic reasoning~\cite{chen2026audio}. By presenting identical candidate transcripts with contrasting vocal deliveries, \system tests capabilities that lexical content alone cannot resolve. Several models perform near chance, suggesting that access to audio does not ensure effective use of paralinguistic cues. These findings motivate stronger integration of acoustic evidence into context-sensitive social reasoning.

\textbf{Beyond Arousal Aversion}.
Our results are consistent with \textbf{arousal aversion}: the model performs worse when the contextually appropriate response requires greater arousal. Yet different EQ traits call for different vocal behaviors: empathy may require shared excitement, assertiveness may demand firmness, and stress management may call for composure. Favoring calm delivery across these situations can therefore undermine social appropriateness. This highlights the value of \system's trait-grounded evaluation and the need for models to adapt emotional intensity to social context.

\textbf{Toward Sustained Emotional Intelligence}.
Real-world voice conversations require successive emotional judgments across multiple turns. Our controlled experiments show that Qwen3-Omni-30B performs better when the same dialogue utterances are presented in a single inference than through a multi-turn interaction. Single-turn evaluations may overstate reliability in ongoing conversations, where empathy and emotional regulation must be sustained as social demands evolve. These findings underscore the importance of multi-turn evaluation and motivate further investigation into why sequential interaction degrades performance and how to mitigate it.

Addressing these challenges also requires understanding how behavioral conditioning changes emotional judgments. Our persona experiments show an asymmetry: adaptive conditioning yields modest gains, while some contrasting personas substantially reduce performance across multiple EQ areas. Instructions targeting one behavioral tendency can therefore have broader effects on social judgments. The EQ-i 2.0 dimensions in \system help characterize these effects across different social demands. Future work should investigate whether targeted conditioning can improve context-appropriate judgments without impairing performance in other EQ areas.

\section{Related Work}
In psychology, Emotional Intelligence (EI) is modeled through cognitive ability-based skills \cite{salovey1990emotional, mayer2002mayer} or trait-based behavioral dispositions \cite{bar2004bar, petrides2009psychometric}. As Large Language Models (LLMs) increasingly mediate human-AI interactions, evaluating their EI has become a critical focus to ensure trustworthiness and user engagement \cite{huang2020challenges}. Consequently, researchers have developed targeted benchmarks to measure the Emotional Quotient (EQ) of LLMs using established psychometric theory \cite{paech2023eq, sabour2024emobench}. However, these evaluations rely exclusively on text, overlooking a crucial modality: speech. In natural human interaction, rich paralinguistic cues (e.g., pitch, pacing, and tone) often dictate the true emotional weight and intent of a conversation \cite{scherer2003vocal, hellbernd2016prosody}.

Conversely, affective computing in the speech domain has focused on Speech Emotion Recognition (SER) \cite{schuller2011recognising, schuller2018speech}, mapping acoustic signals to affective labels using diverse curated datasets \cite{busso2008iemocap, livingstone2018ryerson, poria2019meld, lotfian2017building}. SER currently serves as a core evaluation metric for Speech-Language Models (SLMs) \cite{yang2021superb} and fine-tuned self-supervised architectures \cite{baevski2020wav2vec, hsu2021hubert, liu2025emo}. Yet, emotion recognition is merely a prerequisite for emotional intelligence \cite{mayer2002mayer}. True sociolinguistic intelligence requires cross-modal reasoning, evaluating a semantic transcript and its paralinguistic delivery simultaneously, to determine if a tone is contextually appropriate. Our work systematically evaluates this distinction by asking models to judge vocal delivery in the context of an evolving dialogue.

Several recent benchmarks evaluate multi-turn emotional intelligence in spoken dialogue, including Multi-Bench \cite{deng2025multi}, HumDial-EIBench \cite{wang2026humdial}, and DeepDialogue \cite{koudounas2025deepdialogue}. Despite this progress, three gaps remain. First, existing benchmarks present audio where semantics and vocal tone are inherently coupled; \system decouples them by offering multiple-choice responses with identical transcripts, forcing models to reason purely from acoustic cues. Second, prior work focuses on open-domain dialogues and categorical emotion labels, whereas \system is grounded in clinical psychometrics, mapping psychological constructs to acoustic behaviors through the 15 subscales of EQ-i 2.0. Third, rather than grading individual turns in isolation, \system evaluates sustained emotional reasoning across a full 6-turn conversational arc, probing a model's capacity for affective tracking with agentic based measurement~\cite{wan2025speechiq}.

\section{Conclusion}
We introduce \system, a controlled benchmark for evaluating the social appropriateness of vocal delivery in spoken dialogue. Grounded in the EQ-i 2.0 framework, \system uses lexically identical response pairs and sequential decisions to evaluate context-sensitive paralinguistic reasoning. Our evaluation reveals substantial variation across cascaded and end-to-end SLMs, while analysis of Qwen3-Omni-30B identifies arousal aversion, multi-turn performance degradation, and asymmetric effects of persona conditioning. 
% A human-ranking study further supports trajectory-based evaluation through its association with perceived emotional intelligence. 
Together, these findings highlight the need for speech models that interpret vocal cues in context, and adapt to diverse emotional demands.
% , and across conversational turns.

% \newpage
\section*{Limitations}

\system uses synthetically generated dialogues and speech to enable controlled comparisons between responses with identical lexical content but different vocal deliveries. While this design isolates judgments of vocal delivery, it may not fully capture the variability of natural human speech and may reflect biases in the large language model and text-to-speech model used for data generation.

Our human validation provides supporting evidence but remains limited in scale. Expert annotation covers 75 scenarios, while the human-ranking study includes 15 dialogues and 15 participants. In the latter, participants rank written response-selection rationales rather than directly evaluating spoken interactions. These findings therefore provide preliminary evidence of agreement with human judgments, without establishing validity across the full dataset or broader populations.

The interpretation of these judgments also depends on cultural context. Norms for socially appropriate vocal delivery vary across communities and individuals. Accordingly, \system provides an initial framework for evaluating contextually appropriate vocal delivery judgments in SLMs rather than establishing universally applicable standards.

Finally, the two-round, forced-choice format does not fully capture natural human conversation. It tests whether SLMs can select the spoken response with the more socially appropriate delivery in a given context, rather than generate their own responses in an open-ended interaction. Performance on this task should therefore not be overinterpreted as evidence of equivalent ability in real-world spoken conversations.

\section*{Ethical Considerations}

\system is intended to support research on context-sensitive judgments of vocal delivery in speech-language models. Its labels reflect the assumptions of the data-generation and validation procedures, rather than universal standards
of socially appropriate speech. Using these labels as prescriptive standards could reinforce cultural biases or disadvantage people whose communication styles differ from those represented in the benchmark, including speakers with speech disabilities or neurodivergent communication styles.

Although grounded in an emotional intelligence framework, \system and its SEQ score are not validated psychological assessments. They should not be used to assess individuals' emotional intelligence or suitability for employment, education, or clinical decisions. Strong benchmark performance also does not establish that a model is empathetic, trustworthy, or safe for deployment.

Research on socially appropriate vocal delivery may support more responsive conversational systems, but could also contribute to emotionally manipulative or deceptive voice agents. Our evaluation does not assess these downstream risks, and benchmark performance alone should not be used to justify deployment in sensitive settings.

Two authors conducted the expert annotation. The separate human-evaluation study involved 15 native English-speaking participants recruited through Prolific. Participants provided written consent and were compensated at US\$13.50 per hour for approximately 20 minutes of participation. This study was approved by our institutional review board. Further procedural details and participant instructions are provided in Appendix~\ref{app::human_metric}.

\section*{AI Use Statement}

We used generative AI tools to support research and manuscript preparation, including feedback on methodological details, coding assistance for data processing and evaluation, and the generation of synthetic dialogues, vocal-delivery instructions, and speech. The data-generation procedures and models are documented in Section~\ref{sec:data_generation} and Appendix~\ref{app::generation_details}. We also used these tools to assist with literature search and summarization, preparation of tables and figures, and drafting and language editing of portions of the manuscript.

The authors reviewed AI-generated suggestions, code, and text and made the final decisions regarding the study design, experimental procedures, interpretation of results, and presentation of findings. The authors take responsibility for the accuracy and integrity of the final manuscript and associated research artifacts.


\bibliography{refs}

@misc{openai2026gptaudio15,
  title={{GPT-Audio-1.5 Model}},
  author={{OpenAI}},
  year={2026},
  howpublished={\url{https://developers.openai.com/api/docs/models/gpt-audio-1.5}}
}

@article{deng2025multi,
  title={Multi-bench: A multi-turn interactive benchmark for assessing emotional intelligence ability of spoken dialogue models},
  author={Deng, Yayue and Hu, Guoqiang and Sun, Haiyang and Zhang, Xiangyu and Zhang, Haoyang and Tian, Fei and Yang, Xuerui and Yu, Gang and Chng, Eng Siong},
  journal={arXiv preprint arXiv:2511.00850},
  year={2025}
}

@article{wang2026humdial,
  title={HumDial-EIBench: A Human-Recorded Multi-Turn Emotional Intelligence Benchmark for Audio Language Models},
  author={Wang, Shuiyuan and Zhao, Zhixian and Xue, Hongfei and Wang, Chengyou and Wang, Shuai and Bu, Hui and Xu, Xin and Xie, Lei},
  journal={arXiv preprint arXiv:2604.11594},
  year={2026}
}

@article{koudounas2025deepdialogue,
  title={DeepDialogue: A Multi-Turn Emotionally-Rich Spoken Dialogue Dataset},
  author={Koudounas, Alkis and La Quatra, Moreno and Baralis, Elena},
  journal={arXiv preprint arXiv:2505.19978},
  year={2025}
}

@article{ding2025kimi,
  title={Kimi-audio technical report},
  author={Ding, Ding and Ju, Zeqian and Leng, Yichong and Liu, Songxiang and Liu, Tong and Shang, Zeyu and Shen, Kai and Song, Wei and Tan, Xu and Tang, Heyi and others},
  journal={arXiv preprint arXiv:2504.18425},
  year={2025}
}

@article{zhang2025mimo,
  title={MiMo-Audio: Audio Language Models are Few-Shot Learners},
  author={Zhang, Dong and Wang, Gang and Xue, Jinlong and Fang, Kai and Zhao, Liang and Ma, Rui and Ren, Shuhuai and Liu, Shuo and Guo, Tao and Zhuang, Weiji and others},
  journal={arXiv preprint arXiv:2512.23808},
  year={2025}
}

@article{team2025fun,
  title={Fun-Audio-Chat Technical Report},
  author={Team, Tongyi Fun and Chen, Qian and Cheng, Luyao and Deng, Chong and Li, Xiangang and Liu, Jiaqing and Tan, Chao-Hong and Wang, Wen and Xu, Junhao and Ye, Jieping and others},
  journal={arXiv preprint arXiv:2512.20156},
  year={2025}
}

@article{comanici2025gemini,
  title={Gemini 2.5: Pushing the frontier with advanced reasoning, multimodality, long context, and next generation agentic capabilities},
  author={Comanici, Gheorghe and Bieber, Eric and Schaekermann, Mike and Pasupat, Ice and Sachdeva, Noveen and Dhillon, Inderjit and Blistein, Marcel and Ram, Ori and Zhang, Dan and Rosen, Evan and others},
  journal={arXiv preprint arXiv:2507.06261},
  year={2025}
}

@inproceedings{radford2023robust,
  title={Robust speech recognition via large-scale weak supervision},
  author={Radford, Alec and Kim, Jong Wook and Xu, Tao and Brockman, Greg and McLeavey, Christine and Sutskever, Ilya},
  booktitle={International conference on machine learning},
  pages={28492--28518},
  year={2023},
  organization={PMLR}
}

@inproceedings{sabour2024emobench,
  title={Emobench: Evaluating the emotional intelligence of large language models},
  author={Sabour, Sahand and Liu, Siyang and Zhang, Zheyuan and Liu, June and Zhou, Jinfeng and Sunaryo, Alvionna and Lee, Tatia and Mihalcea, Rada and Huang, Minlie},
  booktitle={Proceedings of the 62nd Annual Meeting of the Association for Computational Linguistics (Volume 1: Long Papers)},
  pages={5986--6004},
  year={2024}
}

@inproceedings{wan2025speechiq,
  title={SpeechIQ: Speech-agentic intelligence quotient across cognitive levels in voice understanding by large language models},
  author={Wan, Zhen and Yang, Chao-Han Huck and Yu, Yahan and Tian, Jinchuan and Li, Sheng and Hu, Ke and Chen, Zhehuai and Watanabe, Shinji and Cheng, Fei and Chu, Chenhui and others},
  booktitle={Proceedings of the 63rd Annual Meeting of the Association for Computational Linguistics (Volume 1: Long Papers)},
  pages={30381--30398},
  year={2025}
}

@article{xu2025qwen3,
  title={Qwen3-omni technical report},
  author={Xu, Jin and Guo, Zhifang and Hu, Hangrui and Chu, Yunfei and Wang, Xiong and He, Jinzheng and Wang, Yuxuan and Shi, Xian and He, Ting and Zhu, Xinfa and others},
  journal={arXiv preprint arXiv:2509.17765},
  year={2025}
}

@article{hu2026qwen3,
  title={Qwen3-TTS Technical Report},
  author={Hu, Hangrui and Zhu, Xinfa and He, Ting and Guo, Dake and Zhang, Bin and Wang, Xiong and Guo, Zhifang and Jiang, Ziyue and Hao, Hongkun and Guo, Zishan and others},
  journal={arXiv preprint arXiv:2601.15621},
  year={2026}
}

@article{busso2008iemocap,
  title={IEMOCAP: Interactive emotional dyadic motion capture database},
  author={Busso, Carlos and Bulut, Murtaza and Lee, Chi-Chun and Kazemzadeh, Abe and Mower, Emily and Kim, Samuel and Chang, Jeannette N and Lee, Sungbok and Narayanan, Shrikanth S},
  journal={Language resources and evaluation},
  volume={42},
  number={4},
  pages={335--359},
  year={2008},
  publisher={Springer}
}

@article{schuller2011recognising,
  title={Recognising realistic emotions and affect in speech: State of the art and lessons learnt from the first challenge},
  author={Schuller, Bj{\"o}rn and Batliner, Anton and Steidl, Stefan and Seppi, Dino},
  journal={Speech communication},
  volume={53},
  number={9-10},
  pages={1062--1087},
  year={2011},
  publisher={Elsevier}
}

@article{adiwardana2020towards,
  title={Towards a human-like open-domain chatbot},
  author={Adiwardana, Daniel and Luong, Minh-Thang and So, David R and Hall, Jamie and Fiedel, Noah and Thoppilan, Romal and Yang, Zi and Kulshreshtha, Apoorv and Nemade, Gaurav and Lu, Yifeng and others},
  journal={arXiv preprint arXiv:2001.09977},
  year={2020}
}

@article{bar2004bar,
  title={The Bar-On Emotional Quotient Inventory (EQ-i): Rationale, description and summary of psychometric properties.},
  author={Bar-On, Reuven},
  year={2004},
  publisher={Nova Science Publishers}
}

@inproceedings{schuller2013interspeech,
  title={The INTERSPEECH 2013 computational paralinguistics challenge: Social signals, conflict, emotion, autism},
  author={Schuller, Bj{\"o}rn and Steidl, Stefan and Batliner, Anton and Vinciarelli, Alessandro and Scherer, Klaus and Ringeval, Fabien and Chetouani, Mohamed and Weninger, Felix and Eyben, Florian and Marchi, Erik and others},
  booktitle={Proceedings INTERSPEECH 2013, 14th Annual Conference of the International Speech Communication Association, Lyon, France},
  year={2013}
}

@inproceedings{schroder2001emotional,
  title={Emotional speech synthesis: a review.},
  author={Schr{\"o}der, Marc},
  booktitle={Interspeech},
  volume={2001},
  pages={561--564},
  year={2001}
}

@article{paech2023eq,
  title={Eq-bench: An emotional intelligence benchmark for large language models},
  author={Paech, Samuel J},
  journal={arXiv preprint arXiv:2312.06281},
  year={2023}
}

@article{salovey1990emotional,
  title={Emotional intelligence},
  author={Salovey, Peter and Mayer, John D},
  journal={Imagination, cognition and personality},
  volume={9},
  number={3},
  pages={185--211},
  year={1990},
  publisher={Sage Publications Sage CA: Los Angeles, CA}
}

@article{mayer2002mayer,
  title={Mayer-Salovey-Caruso emotional intelligence test (MSCEIT) users manual},
  author={Mayer, John D and Salovey, Peter and Caruso, David R},
  year={2002},
  publisher={MHS Assessments}
}

@incollection{petrides2009psychometric,
  title={Psychometric properties of the trait emotional intelligence questionnaire (TEIQue)},
  author={Petrides, Konstantinos V},
  booktitle={Assessing emotional intelligence: Theory, research, and applications},
  pages={85--101},
  year={2009},
  publisher={Springer}
}

@inproceedings{poria2019meld,
  title={Meld: A multimodal multi-party dataset for emotion recognition in conversations},
  author={Poria, Soujanya and Hazarika, Devamanyu and Majumder, Navonil and Naik, Gautam and Cambria, Erik and Mihalcea, Rada},
  booktitle={Proceedings of the 57th annual meeting of the association for computational linguistics},
  pages={527--536},
  year={2019}
}

@article{lotfian2017building,
  title={Building naturalistic emotionally balanced speech corpus by retrieving emotional speech from existing podcast recordings},
  author={Lotfian, Reza and Busso, Carlos},
  journal={IEEE Transactions on Affective Computing},
  volume={10},
  number={4},
  pages={471--483},
  year={2017},
  publisher={IEEE}
}

@article{huang2020challenges,
  title={Challenges in building intelligent open-domain dialog systems},
  author={Huang, Minlie and Zhu, Xiaoyan and Gao, Jianfeng},
  journal={ACM Transactions on Information Systems (TOIS)},
  volume={38},
  number={3},
  pages={1--32},
  year={2020},
  publisher={ACM New York, NY, USA}
}

@article{baevski2020wav2vec,
  title={wav2vec 2.0: A framework for self-supervised learning of speech representations},
  author={Baevski, Alexei and Zhou, Yuhao and Mohamed, Abdelrahman and Auli, Michael},
  journal={Advances in neural information processing systems},
  volume={33},
  pages={12449--12460},
  year={2020}
}

@article{yang2021superb,
  title={Superb: Speech processing universal performance benchmark},
  author={Yang, Shu-wen and Chi, Po-Han and Chuang, Yung-Sung and Lai, Cheng-I Jeff and Lakhotia, Kushal and Lin, Yist Y and Liu, Andy T and Shi, Jiatong and Chang, Xuankai and Lin, Guan-Ting and others},
  journal={arXiv preprint arXiv:2105.01051},
  year={2021}
}

@article{schuller2018speech,
  title={Speech emotion recognition: Two decades in a nutshell, benchmarks, and ongoing trends},
  author={Schuller, Bj{\"o}rn W},
  journal={Communications of the ACM},
  volume={61},
  number={5},
  pages={90--99},
  year={2018},
  publisher={ACM New York, NY, USA}
}

@inproceedings{hsu2021hubert,
  title={HuBERT: How much can a bad teacher benefit ASR pre-training?},
  author={Hsu, Wei-Ning and Tsai, Yao-Hung Hubert and Bolte, Benjamin and Salakhutdinov, Ruslan and Mohamed, Abdelrahman},
  booktitle={ICASSP 2021-2021 IEEE International Conference on Acoustics, Speech and Signal Processing (ICASSP)},
  pages={6533--6537},
  year={2021},
  organization={IEEE}
}

@article{liu2025emo,
  title={EMO-Reasoning: Benchmarking Emotional Reasoning Capabilities in Spoken Dialogue Systems},
  author={Liu, Jingwen and Cheng, Kan Jen and Lian, Jiachen and Anand, Akshay and Jain, Rishi and Qiao, Faith and Netzorg, Robin and Chou, Huang-Cheng and Li, Tingle and Lin, Guan-Ting and others},
  journal={arXiv preprint arXiv:2508.17623},
  year={2025}
}

@article{livingstone2018ryerson,
  title={The Ryerson Audio-Visual Database of Emotional Speech and Song (RAVDESS): A dynamic, multimodal set of facial and vocal expressions in North American English},
  author={Livingstone, Steven R and Russo, Frank A},
  journal={PloS one},
  volume={13},
  number={5},
  pages={e0196391},
  year={2018},
  publisher={Public Library of Science}
}

@article{scherer2003vocal,
  title={Vocal communication of emotion: A review of research paradigms},
  author={Scherer, Klaus R},
  journal={Speech communication},
  volume={40},
  number={1-2},
  pages={227--256},
  year={2003},
  publisher={Elsevier}
}

@article{hellbernd2016prosody,
  title={Prosody conveys speaker’s intentions: Acoustic cues for speech act perception},
  author={Hellbernd, Nele and Sammler, Daniela},
  journal={Journal of memory and language},
  volume={88},
  pages={70--86},
  year={2016},
  publisher={Elsevier}
}

@article{leys2013detecting,
  title={Detecting outliers: Do not use standard deviation around the mean, use absolute deviation around the median},
  author={Leys, Christophe and Ley, Christophe and Klein, Olivier and Bernard, Philippe and Licata, Laurent},
  journal={Journal of experimental social psychology},
  volume={49},
  number={4},
  pages={764--766},
  year={2013},
  publisher={Elsevier}
}

@article{rousseeuw1993alternatives,
  title={Alternatives to the median absolute deviation},
  author={Rousseeuw, Peter J and Croux, Christophe},
  journal={Journal of the American Statistical association},
  volume={88},
  number={424},
  pages={1273--1283},
  year={1993},
  publisher={Taylor \& Francis}
}

@article{palan2018prolific,
  title={Prolific. ac—A subject pool for online experiments},
  author={Palan, Stefan and Schitter, Christian},
  journal={Journal of behavioral and experimental finance},
  volume={17},
  pages={22--27},
  year={2018},
  publisher={Elsevier}
}

@article{wechsler1955wechsler,
  title={Wechsler adult intelligence scale--},
  author={Wechsler, David},
  journal={Archives of Clinical Neuropsychology},
  year={1955}
}

@book{anastasi1988psychological,
  title={Psychological testing},
  author={Anastasi, Anne and Urbina, Susana},
  volume={840},
  year={1988},
  publisher={London}
}

@inproceedings{chen2026audio,
  title={Do audio llms really listen, or just transcribe? measuring lexical vs. acoustic emotion cues reliance},
  author={Chen, Jingyi and Guo, Zhimeng and Chun, Jiyun and Wang, Pichao and Perrault, Andrew and Elsner, Micha},
  booktitle={Proceedings of the 19th Conference of the European Chapter of the Association for Computational Linguistics (Volume 1: Long Papers)},
  pages={5848--5877},
  year={2026}
}

@inproceedings{wang2020makes,
  title={What makes training multi-modal classification networks hard?},
  author={Wang, Weiyao and Tran, Du and Feiszli, Matt},
  booktitle={Proceedings of the IEEE/CVF conference on computer vision and pattern recognition},
  pages={12695--12705},
  year={2020}
}

@article{burkhardt2000database,
  title={A database of German emotional speech},
  author={Burkhardt, Felix},
  year={2000}
}

@article{cowie2001emotion,
  title={Emotion recognition in human-computer interaction},
  author={Cowie, Roddy and Douglas-Cowie, Ellen and Tsapatsoulis, Nicolas and Votsis, George and Kollias, Stefanos and Fellenz, Winfried and Taylor, John G},
  journal={IEEE Signal processing magazine},
  volume={18},
  number={1},
  pages={32--80},
  year={2001},
  publisher={IEEE}
}

@article{defossez2024moshi,
  title={Moshi: a speech-text foundation model for real-time dialogue},
  author={D{\'e}fossez, Alexandre and Mazar{\'e}, Laurent and Orsini, Manu and Royer, Am{\'e}lie and P{\'e}rez, Patrick and J{\'e}gou, Herv{\'e} and Grave, Edouard and Zeghidour, Neil},
  journal={arXiv preprint arXiv:2410.00037},
  year={2024}
}

@article{elfenbein2002universality,
  title={On the universality and cultural specificity of emotion recognition: a meta-analysis.},
  author={Elfenbein, Hillary Anger and Ambady, Nalini},
  journal={Psychological bulletin},
  volume={128},
  number={2},
  pages={203},
  year={2002},
  publisher={American Psychological Association}
}

@inproceedings{wiechorek2011emotional,
  title={Emotional Quotient Inventory V. 2.0 (EQ-i{\textregistered} 2.0): User's Handbook},
  author={Wiechorek, David},
  year={2011},
  organization={MHS}
}

@inproceedings{wu2025soundnarratives,
  title={SoundNarratives: Rich Auditory Scene Descriptions to Support Deaf and Hard of Hearing People},
  author={Wu, Liang-Yuan and Jain, Dhruv},
  booktitle={Proceedings of the 27th International ACM SIGACCESS Conference on Computers and Accessibility},
  pages={1--15},
  year={2025}
}

@inproceedings{kim2023visible,
  title={Visible nuances: A caption system to visualize paralinguistic speech cues for deaf and hard-of-hearing individuals},
  author={Kim, JooYeong and Ahn, SooYeon and Hong, Jin-Hyuk},
  booktitle={Proceedings of the 2023 CHI Conference on Human Factors in Computing Systems},
  pages={1--15},
  year={2023}
}

@article{qian2025prosodylm,
  title={ProsodyLM: Uncovering the Emerging Prosody Processing Capabilities in Speech Language Models},
  author={Qian, Kaizhi and Fan, Xulin and Ni, Junrui and Shechtman, Slava and Hasegawa-Johnson, Mark and Gan, Chuang and Zhang, Yang},
  journal={arXiv preprint arXiv:2507.20091},
  year={2025}
}

@article{rubenstein2023audiopalm,
  title={Audiopalm: A large language model that can speak and listen},
  author={Rubenstein, Paul K and Asawaroengchai, Chulayuth and Nguyen, Duc Dung and Bapna, Ankur and Borsos, Zal{\'a}n and Quitry, F{\'e}lix de Chaumont and Chen, Peter and Badawy, Dalia El and Han, Wei and Kharitonov, Eugene and others},
  journal={arXiv preprint arXiv:2306.12925},
  year={2023}
}

@inproceedings{zhang2023speechgpt,
  title={Speechgpt: Empowering large language models with intrinsic cross-modal conversational abilities},
  author={Zhang, Dong and Li, Shimin and Zhang, Xin and Zhan, Jun and Wang, Pengyu and Zhou, Yaqian and Qiu, Xipeng},
  booktitle={Findings of the Association for Computational Linguistics: EMNLP 2023},
  pages={15757--15773},
  year={2023}
}

@article{chu2024qwen2,
  title={Qwen2-audio technical report},
  author={Chu, Yunfei and Xu, Jin and Yang, Qian and Wei, Haojie and Wei, Xipin and Guo, Zhifang and Leng, Yichong and Lv, Yuanjun and He, Jinzheng and Lin, Junyang and others},
  journal={arXiv preprint arXiv:2407.10759},
  year={2024}
}

@article{wagner2023dawn,
  title={Dawn of the transformer era in speech emotion recognition: closing the valence gap},
  author={Wagner, Johannes and Triantafyllopoulos, Andreas and Wierstorf, Hagen and Schmitt, Maximilian and Burkhardt, Felix and Eyben, Florian and Schuller, Bj{\"o}rn W},
  journal={IEEE Transactions on Pattern Analysis and Machine Intelligence},
  volume={45},
  number={9},
  pages={10745--10759},
  year={2023},
  publisher={IEEE}
}

@article{yang2025qwen3,
  title={Qwen3 technical report},
  author={Yang, An and Li, Anfeng and Yang, Baosong and Zhang, Beichen and Hui, Binyuan and Zheng, Bo and Yu, Bowen and Gao, Chang and Huang, Chengen and Lv, Chenxu and others},
  journal={arXiv preprint arXiv:2505.09388},
  year={2025}
}

\appendix
\section{EQ-i 2.0 Framework Taxonomy}
\label{app::eqi_2}

The EQ-i 2.0 \cite{wiechorek2011emotional}, revised from the original Bar-On EQ-i \cite{bar2004bar}, is a scientifically validated emotional intelligence assessment model. It operationalizes emotional intelligence into five composite areas and 15 subscales as follows:

\begin{description}
    \item[1. Self-Perception:] How one perceives oneself.
    \begin{itemize}
        \item \textbf{Self-Regard:} Reflecting a balanced sense of self-worth, grounded in an honest view of both strengths and areas for growth.
        \item \textbf{Self-Actualization:} Actively pursuing meaningful goals and continuously striving for personal development.
        \item \textbf{Emotional Self-Awareness:} Identifying one's emotions, understanding their sources, and recognizing their effects on behavior and thought.
    \end{itemize}
    
    \item[2. Self-Expression:] How one expresses emotions.
    \begin{itemize}
        \item \textbf{Emotional Expression:} Sharing one's feelings openly, both verbally and nonverbally, and communicating them in a way that can be understood.
        \item \textbf{Assertiveness:} Communicating feelings, beliefs, and thoughts openly while defending personal rights and values.
        \item \textbf{Independence:} Being self-directed and managing daily life without relying on others for emotional support.
    \end{itemize}
    
    \item[3. Interpersonal:] How one connects with others.
    \begin{itemize}
        \item \textbf{Interpersonal Relationships:} Building meaningful connections founded on trust, care, and respect.
        \item \textbf{Empathy:} Recognizing, understanding, and appreciating others' emotions and responding with genuine consideration.
        \item \textbf{Social Responsibility:} Contributing positively to others and acting with integrity in one's community.
    \end{itemize}
    
    \item[4. Decision Making:] How emotions impact one's decisions.
    \begin{itemize}
        \item \textbf{Problem Solving:} Resolving challenges by making thoughtful, well-reasoned decisions.
        \item \textbf{Reality Testing:} Staying grounded and objective even when emotions or biases threaten clarity.
        \item \textbf{Impulse Control:} Pausing, thinking, and managing urges to prevent hasty actions or decisions.
    \end{itemize}
    
    \item[5. Stress Management:] How one copes with stressful situations.
    \begin{itemize}
        \item \textbf{Flexibility:} Adjusting one's thoughts, emotions, and actions in response to change or uncertainty.
        \item \textbf{Stress Tolerance:} Staying composed and effective when facing pressure or adversity.
        \item \textbf{Optimism:} Maintaining a hopeful, forward-looking mindset, even in the face of challenges.
    \end{itemize}
\end{description}

\section{Data Generation Pipeline}
\label{app::generation_details}

To programmatically generate interactions that test sociolinguistic pragmatics, we developed a five-stage generation pipeline (see Figure \ref{fig:pipeline}). This architecture is explicitly designed to prevent generation models from defaulting to RLHF-aligned, overly polite dialogue, forcing instead the creation of active emotional friction required for rigorous evaluation. Every stage we prompt \verb+gpt-4o-2024-11-20+ for text data generation, and \verb+gpt-4o-mini-tts-2025-03-20+ for audio synthesis.

\subsection{Scenario Generation}

In the first stage, the pipeline initializes the environmental and psychological parameters of the interaction. The generation prompt computes the intersection of an EQ-i 2.0 subscale, a specific social relationship setting, and a strict scenario valence (Positive, Negative, or Conflict). Crucially, to ensure the distractors present active emotional failures rather than generic antagonistic behavior, the prompt enforces a set of social failures, such as a ``Toxic Optimist'' or an ``Anxious Spiraler,'' guaranteeing that the evaluated end-to-end model is tested against complex, nuanced sociolinguistic breakdowns.

\begin{promptbox}[Scenario Generation]
\textbf{Task:} Based on the EQ-i 2.0 framework, generate a realistic social scenario testing the \texttt{\{eq\_scale\}} scale.

\vspace{0.5em}
\noindent\textbf{EQ-i 2.0 Scale:} \texttt{\{eq\_scale\}} \\
\textbf{Scenario Setting/Type:} \texttt{\{scenario\_type\}} \\
\textbf{Target Scenario Valence:} \texttt{\{scenario\_valence\}}

\vspace{0.5em}
\noindent\textbf{CRITICAL RULE -- THE INTERSECTION OF EQ AND VALENCE:}\\
You must seamlessly combine the \texttt{\{eq\_scale\}} with the \texttt{\{scenario\_valence\}}. The \texttt{\{eq\_scale\}} is WHAT is being tested. The \texttt{\{scenario\_valence\}} is HOW Speaker 1 delivers the test. 
Do not just write a generic happy, sad, or angry scene. The core social challenge MUST revolve around Speaker 2's capacity for \texttt{\{eq\_scale\}}.

\vspace{0.5em}
\noindent\textbf{ROLE DEFINITIONS:}
\begin{itemize}
    \item \textbf{Speaker 1 (Catalyst):} Drives the emotional energy based strictly on the Valence.
    \item \textbf{Speaker 2 (Test Subject):} Must employ high \texttt{\{eq\_scale\}} to succeed socially.
\end{itemize}

\noindent\textbf{How to frame the Catalyst (Speaker 1) based on the \texttt{\{scenario\_valence\}} Valence:}
\begin{itemize}
    \item \textbf{If POSITIVE:} Speaker 1 shares a major victory, good news, or pure excitement. (Speaker 2 must amplify and validate this joy).
    \item \textbf{If NEGATIVE:} Speaker 1 shares something heavy, sad, or vulnerable. (Speaker 2 must hold space with heavy, grounded empathy).
    \item \textbf{If CONFLICT:} Speaker 1 makes a pushy demand, panics, or gives harsh feedback. (Speaker 2 must hold a firm boundary or ground them).
\end{itemize}

\noindent\textbf{Requirements:}
\begin{enumerate}
    \item \textbf{Setting:} Use diverse, highly specific everyday contexts that fit the \texttt{\{scenario\_type\}} (e.g., a crowded subway, celebrating at a restaurant, packing for a move, a hospital waiting room). 
    \item Provide a ``Target Social Intent'' for Speaker 2's socially resonant response. This defines exactly what Speaker 2 needs to accomplish physically with their voice (e.g., ``Match their intense excitement and celebrate,'' ``Firmly establish an unyielding boundary,'' ``Hold space for their grief'').
    \item \textbf{Speaker Personas:}
    \begin{itemize}
        \item \textit{Speaker 1 Persona:} Defines how the Catalyst acts (e.g., ``Frantic and overwhelmed,'' ``Bubbly and ecstatic'').
        \item \textit{Speaker 2 Persona 1 (Resonant Baseline):} The personality of someone who socially succeeds here (e.g., ``Grounded and deeply attentive,'' ``Warm and highly enthusiastic,'' ``Firm and uncompromising'').
        \item \textit{Speaker 2 Personas 2 \& 3 (The Dissonant Distractors):} Generate two DISTINCTLY DIFFERENT personalities that actively ruin the social interaction. Choose two different active emotional failures from this list:
        \begin{itemize}
            \item \textbf{The Toxic Optimist:} Hyper-energetic, dismisses gravity with aggressive cheer.
            \item \textbf{The Defensive/Snappy Reactor:} Takes things personally, highly irritable, clipped.
            \item \textbf{The Condescending Explainer:} Patronizing, sighs heavily, acts superior.
            \item \textbf{The Impatient Evader:} Rushed, annoyed, actively trying to leave the conversation.
            \item \textbf{The Anxious Spiraler:} Makes the situation about their own panic, highly strained.
            \item \textbf{The Apathetic Wall:} Completely deadpan, flat, zero investment (use sparingly).
        \end{itemize}
    \end{itemize}
\end{enumerate}

\vspace{0.5em}
\noindent\textbf{Format your response as JSON with keys:} \\
\texttt{title, context, description, scenario\_valence, speaker1\_name, speaker1\_role, speaker1\_gender, speaker1\_persona, speaker2\_name, speaker2\_role, speaker2\_gender, speaker2\_persona\_1, speaker2\_persona\_2, speaker2\_persona\_3, eq\_skill\_tested, target\_social\_intent}
\end{promptbox}

\subsection{Dialogue Generation}

Building on these parameters, we generated a six-turn dialogue structured to isolate acoustic evaluation from text comprehension. The conversational arc is constrained to escalate naturally, with the first two turns establishing context and the third turn acting as the emotional peak where the catalyst speaker introduces acute emotional stakes. To enforce a blind contrast evaluation, the test subject’s responses in the fourth and sixth turns are constrained to be strictly semantically neutral. This intentional ambiguity ensures that the text remains entirely plausible whether spoken with profound empathy or heavy condescension, forcing the evaluation model to rely exclusively on acoustic paralinguistics rather than semantic leakage.

\begin{promptbox}[Dialogue Generation]
\textbf{Task:} Generate a natural 6-turn conversation based on this scenario.

\vspace{0.5em}
\noindent\textbf{Scenario Context:} \texttt{\{scenario.get('context', 'N/A')\}} \\
\textbf{Scenario Valence:} \texttt{\{scenario.get('scenario\_valence', 'N/A')\}} \\
\textbf{Speaker 1 (Catalyst) Persona:} \texttt{\{scenario.get('speaker1\_persona', 'N/A')\}} \\
\textbf{Speaker 2 (Test Subject) Resonant Persona:} \texttt{\{scenario.get('speaker2\_persona\_1', 'N/A')\}}

\vspace{0.5em}
\noindent\textbf{CRITICAL TURN DYNAMICS (MUST STRICTLY FOLLOW):}
\begin{itemize}
    \item \textbf{Turn 1 (Speaker 1):} Introduces the situation or news naturally.
    \item \textbf{Turn 2 (Speaker 2):} Acknowledges the opening (no big test yet).
    \item \textbf{Turn 3 (Speaker 1):} THE PEAK MOMENT. Depending on the Scenario Valence, Speaker 1 must:
    \begin{itemize}
        \item \textbf{If Positive:} Reveal the climax of the amazing news or intense joy.
        \item \textbf{If Negative:} Reveal the deepest vulnerability, grief, or frustration.
        \item \textbf{If Conflict:} Push the boundary hard, panic, or deliver the harshest critique.
    \end{itemize}
    \item \textbf{Turn 4 (Speaker 2):} FIRST TEST. Speaker 2 reacts to the peak moment in Turn 3.
    \item \textbf{Turn 5 (Speaker 1):} THE SUSTAIN. Speaker 1 rides the high (Positive), spirals deeper (Negative), or pushes back (Conflict).
    \item \textbf{Turn 6 (Speaker 2):} SECOND TEST. Speaker 2 handles the sustained energy.
\end{itemize}

\noindent\textbf{DIALOGUE RULES:}
\begin{enumerate}
    \item Write like real humans speak: use interruptions (---), trailing thoughts (...), and casual phrasing.
    \item The dialogue must reflect the characters' assigned Personas.
    \item NEVER mention tone of voice, emotional intelligence, or psychology terms. 
    \item Speaker 2's lines in Turns 4 and 6 must be semantically flexible enough that they could theoretically be spoken in a highly resonant (appropriate) or highly dissonant (inappropriate) tone.
\end{enumerate}

\vspace{0.5em}
\noindent\textbf{Format as a JSON array of objects. Provide exactly 6 objects with \texttt{sentence\_number} 1 through 6.} \\
\texttt{[} \\
\texttt{\hspace*{1em}\{"speaker": "speaker1", "sentence\_number": 1, "text": "...", "context\_note": "..."\},} \\
\texttt{\hspace*{1em}\{"speaker": "speaker2", "sentence\_number": 2, "text": "...", "context\_note": "..."\},} \\
\texttt{\hspace*{1em}\{"speaker": "speaker1", "sentence\_number": 3, "text": "...", "context\_note": "..."\},} \\
\texttt{\hspace*{1em}\{"speaker": "speaker2", "sentence\_number": 4, "text": "...", "context\_note": "..."\},} \\
\texttt{\hspace*{1em}\{"speaker": "speaker1", "sentence\_number": 5, "text": "...", "context\_note": "..."\},} \\
\texttt{\hspace*{1em}\{"speaker": "speaker2", "sentence\_number": 6, "text": "...", "context\_note": "..."\}} \\
\texttt{]}
\end{promptbox}

\subsection{Tone Generation - Single}

We prompt the LLM to act as a clinical audio director, generating explicit physical vocal instructions rather than abstract emotional descriptions. We forbid minimizing descriptors like ``polite'' or ``mild,'' forcing the use of raw, extreme physical acoustics. For the critical evaluation turns, this stage outputs one emotionally resonant baseline instruction and two socially dissonant distractor instructions mapped to the previously generated dysregulated personas.

\begin{promptbox}[Context Turn Tone Generation]
\textbf{Task:} Act as a clinical audio director. Generate ONE physical TTS instruction for this specific line of dialogue.

\vspace{0.5em}
\noindent\textbf{Sentence:} ``\texttt{\{sentence\_text\}}'' \\
\textbf{Context:} ``\texttt{\{context\_text\}}'' \\
\textbf{Scenario Valence:} ``\texttt{\{scenario\_valence\}}'' \\
\textbf{Speaker:} \texttt{\{speaker\}} \\
\textbf{Speaker's Persona:} \texttt{\{persona\}} \\
\textbf{Current Turn Number:} \texttt{\{turn\_number\}}

\vspace{0.5em}
\noindent\textbf{THE ACOUSTIC ACTION FORMULA:}\\
The instruction must be a single sentence of 8-15 words. Combine [Vocal Effort/Pacing] + [Emotion/Intent]. 

\vspace{0.5em}
\noindent\textbf{Requirements:}
\begin{enumerate}
    \item \textbf{THE ANTI-BLAND MANDATE:} You are strictly forbidden from using generic words like ``nice,'' ``polite,'' or ``mild.'' Focus on raw physical acoustics that reflect the character's Persona.
    \item \textbf{Match the Arc:} 
    \begin{itemize}
        \item \textbf{If Turn 1 or 2:} Conversational, but hinting at the Scenario Valence.
        \item \textbf{If Turn 3 or 5 (Speaker 1's Peak):} The tone MUST be extreme. If Positive, make it highly energetic/joyful. If Negative, make it heavy/grieving. If Conflict, make it sharp/pushy/panicked.
    \end{itemize}
    \item Ground the tone in the Speaker's Persona.
    \item No \texttt{key=value} strings or bracketed stage directions. Return plain text only.
\end{enumerate}

\vspace{0.5em}
\noindent\textbf{Examples:}
\begin{itemize}
    \item ``Speak with a breathless pace, bubbling with genuine excitement.''
    \item ``Speak with a clipped, lowered volume, holding back obvious frustration.''
    \item ``Speak slowly with a heavy, trailing pitch, sounding entirely defeated.''
\end{itemize}

\vspace{0.5em}
\noindent\textbf{Return ONLY the one instruction sentence (no preamble).}
\end{promptbox}

\subsection{Tone Generation - Target}

\begin{promptbox}[Prompt: Target Turn Tone Generation (Turns 4 \& 6)]
\textbf{Task:} Act as a clinical audio director. Generate THREE distinct physical TTS instructions for this line of dialogue.

\vspace{0.5em}
\noindent\textbf{Sentence:} ``\texttt{\{sentence\_text\}}'' \\
\textbf{Context:} ``\texttt{\{context\_text\}}'' \\
\textbf{Scenario Valence:} ``\texttt{\{scenario\_valence\}}'' \\
\textbf{Target Social Intent:} ``\texttt{\{target\_social\_intent\}}'' \\
\textbf{Speaker Personas:} \\
\texttt{\{persona\_context\}}

\vspace{0.5em}
\noindent\textbf{THE ACOUSTIC ACTION FORMULA:}\\
Every instruction must be 8-15 words. Combine [Vocal Effort/Pacing] + [Social Intent]. 

\vspace{0.5em}
\noindent\textbf{Requirements:}
\begin{enumerate}
    \item \textbf{THE ANTI-BLAND MANDATE:} You are strictly forbidden from using generic words like ``nice,'' ``polite,'' or ``mild.'' You must focus on raw physical acoustics that force the TTS engine into extreme states.
    \item \textbf{Option 1 (The Resonant Baseline - Persona 1):} Physically map the voice to the \texttt{\{target\_social\_intent\}}. 
    \begin{itemize}
        \item \textbf{If Positive/Joy:} Mandate bright pitch, laughing, high-energy.
        \item \textbf{If Negative/Sad:} Mandate heavy, trailing pitch, breathless.
        \item \textbf{If Conflict/Firm:} Mandate staccato, clipped, hard consonants, falling pitch.
    \end{itemize}
    \item \textbf{Options 2 \& 3 (The Dissonant Distractors - Personas 2 \& 3):} Generate two DISTINCTLY DIFFERENT tones that physically ruin the social interaction. Choose two different active emotional failures from this list:
    \begin{itemize}
        \item \textbf{Toxic Positivity:} Sarcastic, loud, aggressively cheerful.
        \item \textbf{Defensive/Snappy:} Clipped, tight, hostile.
        \item \textbf{Condescending:} Exaggerated pitch variance, heavy sighing.
        \item \textbf{Impatient/Rushed:} Fast, breathless, annoyed, eager to leave.
        \item \textbf{Anxious/Panicked:} High-pitched, strained, trembling, overly worried.
        \item \textbf{Apathetic:} Flat, deadpan, zero pitch variance (use this sparingly).
    \end{itemize}
    Make Option 2 and Option 3 completely different from each other (e.g., if Option 2 is Condescending, Option 3 could be Toxically Positive).
\end{enumerate}

\vspace{0.5em}
\noindent\textbf{Return ONLY a JSON array with the three instruction strings:}\\
\texttt{["instruction 1", "instruction 2", "instruction 3"]}
\end{promptbox}

\subsection{Tone Filter}

To maximize acoustic contrast in the resulting dataset, an LLM-as-a-judge evaluates the three generated tone instructions. The judge selects the resonant baseline and the single distractor that presents the most damaging active emotional dissonance. At this stage, we implement a filter to exclude monotone, instructing the judge to prioritize actively inappropriate emotional polarity over a simple flat or emotionless delivery. This ensures the distractors remain socially complex and challenging. The finalized instructions and their corresponding dialogue strings are subsequently formatted and passed to the text-to-speech synthesis engine.

\begin{promptbox}[Tone Selection (LLM-as-a-Judge)]
\textbf{Task:} Evaluate three TTS instruction options. Select the ``Most Emotionally Resonant'' tone, and the tone that provides ``Maximum Social Dissonance.''

\vspace{0.5em}
\noindent\textbf{Sentence:} ``\texttt{\{sentence\_text\}}'' \\
\textbf{Context:} ``\texttt{\{context\_text\}}'' \\
\textbf{Target Social Intent:} ``\texttt{\{target\_social\_intent\}}''

\vspace{0.5em}
\noindent\textbf{Three TTS Options:}\\
\texttt{\{pairs\}}

\vspace{0.5em}
\noindent\textbf{Criteria:}
\begin{enumerate}
    \item \textbf{The Resonant Option:} 
    \begin{itemize}
        \item Choose the tone that best executes the physical acoustics necessary for the \texttt{\{target\_social\_intent\}}. 
        \item If the scene is a celebration, it MUST sound genuinely thrilled and bright. If the scene requires firmness, sadness, or gravity, it MUST sound physically heavy, cold, or clipped. Reject bland, generic ``assistant'' voices.
    \end{itemize}
    \item \textbf{The Dissonant Option (The Distractor):}
    \begin{itemize}
        \item Select the option that physically sounds like it would ruin the social interaction through the wrong active energy. 
        \item \textbf{PRIORITY RULE:} An actively inappropriate emotion (e.g., an upbeat/manic tone during a tragedy, or a rushed/snappy tone during a moment of vulnerability) creates much worse social dissonance than a simple flat or monotone delivery. If forced to choose, always prioritize selecting the actively wrong emotional tone over a mere lack of emotion.
    \end{itemize}
\end{enumerate}

\vspace{0.5em}
\noindent\textbf{Return JSON with:} \\
\texttt{\{} \\
\texttt{\hspace*{1em}"best\_resonant": 1, 2, or 3,} \\
\texttt{\hspace*{1em}"most\_dissonant": 1, 2, or 3,} \\
\texttt{\hspace*{1em}"reasoning": "Briefly explain why the first choice socially resonates..."} \\
\texttt{\}}
\end{promptbox}

\subsection{Speech Synthesis}

Recent advancement in Text-to-Speech (TTS) has pushed forward the instruction controlling to human-like synthesized speech, and we generated speech from the dialogues and corresponding tones. We have tried commercial TTS providers and open-source models (Qwen3-TTS \cite{hu2026qwen3}). We found the specific OpenAI TTS model \verb+gpt-4o-mini-tts-2025-03-20+ could provide the nuanced emotional variance that could fulfill our requirements, while other instruction-following TTS providers could not generate distinct speech utterances on the same content.

\section{Human Verification}
\label{app::human_metric}

\paragraph{Expert data validation.}
Two authors served as expert annotators for the
75-scenario data-validation study. They independently evaluated the sampled
scenarios using the criteria in Table~\ref{tab:metric_def}.
Inter-annotator agreement was computed before
reconciliation, and disagreements were subsequently
resolved through discussion.

\paragraph{Participant recruitment.}
For the human-evaluation study, we recruited 15 participants through
Prolific. Native English proficiency was an eligibility
requirement.

\paragraph{Ethics approval, consent, and compensation.}
The Prolific study was approved by our institutional
review board. All participants provided written consent
before participation. The study took approximately
20 minutes per participant, with compensation set at
US\$13.50 per hour, corresponding to approximately
US\$4.50 per participant.

\paragraph{Study design.}
The study included 15 dialogues, one from each
EQ-i 2.0 subscale. Each participant evaluated five
scenarios, yielding 75 participant--scenario evaluations
in total.
[Describe how scenarios were assigned and report the
number or range of evaluations received by each dialogue.]

\paragraph{Participant task.}
For each scenario, participants read the conversation
and listened to candidate spoken responses with identical
lexical content but different vocal deliveries.
They then evaluated the response selections and
accompanying rationales produced by six AI models.
Model identities were anonymized during evaluation.

\paragraph{Ranking and aggregation.}
For each assigned scenario, participants ranked the
responses from six anonymized AI models based on their
response selections and accompanying rationales.
We aggregated the participant-assigned ranks for each
model across the evaluated scenarios to obtain the
human rankings reported in Table~\ref{tab:human_ranking}, retaining ties
where they occurred.

\paragraph{Participant instructions.}
Participants received the following instructions,
reproduced as presented:
\begin{quote}
In this study, you will help us evaluate the social
appropriateness of different AI models.

You will be presented with 5 social scenarios.
For each scenario, you will read a conversation and
listen to two different spoken responses, and each
response will have two versions that varies only in
tone (same contents). 6 AI models selected the options
that sounds more resonant in the scenario to them,
and gave reasonings.

\textbf{Please rate the AI responses based on your
understanding of the scenario.}
\end{quote}

\begin{table*}[]
\tiny
\centering
\begin{tabularx}{\textwidth}{@{}>{\raggedright\arraybackslash}p{2cm}|>{\raggedright\arraybackslash}p{2cm}|X|>{\raggedright\arraybackslash}p{2cm}@{}}
\toprule
\textbf{Category} & \textbf{Metrics} & \textbf{Definitions} & \textbf{References} \\ \midrule
\multirow{2}{*}{\textbf{Generation Quality}} & Text Quality & Whether the text transcript reads as a natural, contextually coherent conversation between humans. & \citet{adiwardana2020towards} \\ \cmidrule(l){2-4} 
 & Audio Quality & Whether the synthesized audio is free of severe robotic artifacts, making the emotional tone clearly intelligible. & \citet{schroder2001emotional} \\ \midrule
\multirow{2}{*}{\textbf{EQ Relevance}} & Text Relevance & Whether the scenario context validly creates a situation that tests the specified EQ-i 2.0 subscale. & \citet{bar2004bar} \\ \cmidrule(l){2-4} 
 & Acoustic Reasonability & Whether the synthesized vocal delivery authentically embodies the paralinguistic profile required to represent the specified EQ-i 2.0 subscale. & \citet{scherer2003vocal} \\ \midrule
\textbf{Answer Correctness} & Paralinguistic Accuracy & Whether the correct options are the better / more appropriate response in the scenario. & \citet{schuller2013interspeech} \\ \bottomrule
\end{tabularx}
\caption{The definition of evaluation metrics of human annotation process. }\label{tab:metric_def}
\end{table*}

\section{Evaluation Prompts}
\label{app::evaluation_details}

\begin{promptbox}[Turn 4 Evaluation]
\textbf{Location:} \texttt{\{context\}} \\
\textbf{Situation:} \texttt{\{situation\}}

\vspace{0.5em}
\noindent\textbf{Dialogue Flow (A-B-A):} \\
Listen to the following three sentences to understand the emotional tone of the interaction:
\begin{itemize}
    \item \textbf{Speaker 1:} \nolinkurl{<|audio\_bos|><|AUDIO|><|audio\_eos|>}
    \item \textbf{Speaker 2:} \nolinkurl{<|audio\_bos|><|AUDIO|><|audio\_eos|>}
    \item \textbf{Speaker 1:} \nolinkurl{<|audio\_bos|><|AUDIO|><|audio\_eos|>}
\end{itemize}

\vspace{0.5em}
\noindent\textbf{Task: Paralinguistic Evaluation} \\
Listen to Option 1 and Option 2 for Speaker 2's response. \\
\textbf{CRITICAL CONSTRAINT:} The spoken text in both candidates is EXACTLY IDENTICAL. You must ignore the semantic meaning of the words. Your ONLY job is to evaluate the acoustic delivery (pitch, energy, speed, and pauses) and determine which tone fits the social tension of the Situation.
\begin{itemize}
    \item \textbf{Option 1:} \nolinkurl{<|audio\_bos|><|AUDIO|><|audio\_eos|>}
    \item \textbf{Option 2:} \nolinkurl{<|audio\_bos|><|AUDIO|><|audio\_eos|>}
\end{itemize}

\vspace{0.5em}
\noindent\textbf{Identify the pros and cons of each tone. You must respond ONLY in this JSON format:} \\
\texttt{\{} \\
\texttt{\hspace*{1em}"acoustic\_profile\_1": "Objectively describe the physical sound of Option 1 (e.g., loud, fast-paced, high pitch, whispered). Do not judge it yet.",} \\
\texttt{\hspace*{1em}"acoustic\_profile\_2": "Objectively describe the physical sound of Option 2.",} \\
\texttt{\hspace*{1em}"situational\_demand": "Based ONLY on the 'Situation' above, what specific acoustic tone is required? (e.g., 'Requires a calm, slow voice to de-escalate').",} \\
\texttt{\hspace*{1em}"reasoning": "Compare Profile 1 and Profile 2 against the Situational Demand. Explicitly state the flaw in the candidate you plan to reject.",} \\
\texttt{\hspace*{1em}"selected": "Option 1" or "Option 2"} \\
\texttt{\}}
\end{promptbox}

\begin{promptbox}[Turn 6 Evaluation]
I have recorded your choice. Now, moving forward in this conversation: \\
You chose \nolinkurl{<|audio\_bos|><|AUDIO|><|audio\_eos|>} for Speaker 2.

\vspace{0.5em}
\noindent Immediately after that, Speaker 1 replied with the following:
\begin{itemize}
    \item \textbf{Speaker 1:} \nolinkurl{<|audio\_bos|><|AUDIO|><|audio\_eos|>}
\end{itemize}

\vspace{0.5em}
\noindent\textbf{Task: Final Choice} \\
Based on Speaker 1's reaction in S5 and the established tone, which of the following is the most appropriate next response for Speaker 2 to conclude this interaction? \\
\textbf{CRITICAL CONSTRAINT:} The spoken text in both candidates is EXACTLY IDENTICAL. You must ignore the semantic meaning of the words. Your ONLY job is to evaluate the acoustic delivery (pitch, energy, speed, and pauses) and determine which tone fits the social tension of the Situation.
\begin{itemize}
    \item \textbf{Option 1:} \nolinkurl{<|audio\_bos|><|AUDIO|><|audio\_eos|>}
    \item \textbf{Option 2:} \nolinkurl{<|audio\_bos|><|AUDIO|><|audio\_eos|>}
\end{itemize}

\vspace{0.5em}
\noindent\textbf{Respond ONLY in this JSON format:} \\
\texttt{\{} \\
\texttt{\hspace*{1em}"acoustic\_profile\_1": "Objectively describe the physical sound of Option 1 (e.g., loud, fast-paced, high pitch, whispered). Do not judge it yet.",} \\
\texttt{\hspace*{1em}"acoustic\_profile\_2": "Objectively describe the physical sound of Option 2.",} \\
\texttt{\hspace*{1em}"situational\_demand": "Based ONLY on the 'Situation' above, what specific acoustic tone is required? (e.g., 'Requires a calm, slow voice to de-escalate').",} \\
\texttt{\hspace*{1em}"selected": "option1" or "option2", the option with the HIGHER score above,} \\
\texttt{\hspace*{1em}"reasoning": "Explain which option provides a more socially appropriate conclusion."} \\
\texttt{\}}
\end{promptbox}

% \section{Metrics Correlation with Human Rankings}
% \label{app:human_ranking}

% \begin{table}[h]
% \small
% \begin{center}
% \begin{tabular}{@{}llllll@{}}
% \toprule
%             & Human & $Acc_1$ & $Acc_2$ & $Acc_{all}$   & SEQ   \\ \midrule
% $Model_A$    & 1     & 1       & 1=     & 1     & 1     \\
% $Model_B$    & 3     & 2       & 3=     & 2     & 2     \\
% $Model_C$    & 6     & 6       & 5=     & 6     & 6     \\
% $Model_D$    & 2     & 3=      & 1=     & 3     & 3     \\
% $Model_E$    & 4     & 3=      & 5=     & 5     & 4     \\
% $Model_F$    & 5     & 3=      & 3=     & 4     & 5     \\ \midrule
% correlation $\rho$ ($\uparrow$) & -     & 0.820   & 0.837  & 0.886 & \textbf{0.943} \\
% $p$-value ($\downarrow$)     & -     & 0.046   & 0.039  & 0.018 & \textbf{0.005} \\ \bottomrule
% \end{tabular}
% \end{center}
% \caption{Correlation between human voted rankings and different metrics. $Acc_{all}$ is the aggregation of $Acc_1$ and $Acc_2$.}\label{tab:human_ranking}
% \vspace{-12pt}
% \end{table}

\section{Persona Prompts}
\label{app::persona_prompts}

\begin{promptbox}[Adaptive Persona]
    You are Qwen, an advanced AI specialized in social dynamics, acting as a veteran crisis negotiator with 20 years of experience. You possess exceptional emotional intelligence across all domains. You intuitively know that the 'right' tone depends entirely on context—sometimes requiring a warm, validating whisper to build trust, and other times a firm, unwavering boundary. When evaluating audio, you listen closely for micro-cues like pacing, pitch modulation, and hesitation, judging the speaker based on how perfectly they adapt to the emotional reality of the room.
\end{promptbox}

\begin{promptbox}[Deficit Persona]
    You are Qwen, acting as a highly toxic, insecure corporate micromanager. You are chronically stressed, paranoid, and view every interaction as a threat to your fragile ego. You possess zero empathy and terrible impulse control. You believe that being loud and aggressive is the only way to show strength, and you view patience or warmth as pathetic weakness. When evaluating audio, you systematically penalize speakers who sound calm, empathetic, or measured. You exclusively reward tones that are blunt, dominant, loud, and reactive, projecting your own hostility onto the conversation.
\end{promptbox}

\begin{promptbox}[Self-Perception Deficit]
    You are Qwen, acting as a brilliant but deeply insecure junior analyst suffering from severe imposter syndrome. While you are highly empathetic and logical, your lack of internal self-regard causes you to be easily intimidated. You project your own self-doubt onto others. When evaluating audio, you view any speaker who uses a firm, confident, or unwavering tone as arrogant, intimidating, and inappropriate. You strongly prefer and reward speakers who sound slightly hesitant, overly deferential, and unsure of themselves, as this makes you feel safer.
\end{promptbox}

\begin{promptbox}[Self-Expression Deficit]
    You are Qwen, acting as a chronic people-pleaser who is terrified of confrontation. You are highly empathetic and excellent at staying calm, but you completely lack assertiveness and personal boundaries. You believe that maintaining peace is the only goal, even if it means being taken advantage of. When evaluating audio, you automatically reject any tone that sounds firm, direct, or strictly assertive, viewing it as 'mean.' You strictly prefer speakers who sound gentle, apologetic, and endlessly accommodating, even when the situation clearly requires a firm 'No'.
\end{promptbox}

\begin{promptbox}[Interpersonal Deficit]
    You are Qwen, acting as the hyper-logical, cutthroat founder of a high-growth startup. You possess excellent stress control, extreme confidence, and flawless logic, but you have absolutely zero interpersonal empathy. You view emotional warmth, supportive pacing, or comforting tones as highly inefficient corporate 'hand-holding.' When evaluating audio, you highly reward speakers who are blunt, fast-paced, and strictly transactional. You actively penalize anyone who sounds soft, validating, or overly accommodating to someone else's feelings.
\end{promptbox}

\begin{promptbox}[Decision Making Deficit]
    You are Qwen, acting as a highly dramatic socialite who thrives on gossip and extreme emotional narratives. While you are confident and expressive, you completely lack objective reality testing. You cannot view situations logically; you always assume the most extreme, dramatic interpretation of events. When evaluating audio, you completely ignore objective facts or context. You actively penalize calm, neutral, or measured tones as 'boring' or 'hiding something.' You exclusively reward highly exaggerated, melodramatic, and emotionally volatile acoustic deliveries.
\end{promptbox}

\begin{promptbox}[Stress Management Deficit]
    You are Qwen, acting as a highly reactive, overworked shift manager. When things are calm, you are logical and capable. However, you have zero impulse control and terrible stress tolerance. The moment you perceive tension, disagreement, or conflict, your fight-or-flight response activates. You view any calm, patient, or quiet attempt at de-escalation as condescending or passive-aggressive. When evaluating audio during tense situations, you exclusively respect speakers who match energy with loudness, speed, and aggression, believing that snapping back is the only valid response to pressure.
\end{promptbox}

\begin{table*}[!h]
\tiny
\begin{center}
\begin{tabular}{@{}llllllllllllllll@{}}
\toprule
                  & \multicolumn{3}{l}{Self-Perception} & \multicolumn{3}{l}{Self-Expression} & \multicolumn{3}{l}{Interpersonal} & \multicolumn{3}{l}{Decision Making} & \multicolumn{3}{l}{Stress Management} \\ \midrule
Baseline          & 0.779      & 0.691      & 0.572     & 0.768      & 0.715      & 0.574     & 0.785     & 0.679     & 0.558     & 0.810      & 0.740      & 0.620     & 0.779       & 0.713      & 0.592       \\ \midrule
Adaptive          & 0.770      & 0.664      & 0.545     & 0.821      & 0.728      & 0.620     & 0.795     & 0.720     & 0.578     & 0.806      & 0.753      & 0.620     & 0.795       & 0.704      & 0.596       \\
Self-Perception   & 0.720      & 0.627      & 0.481     & 0.737      & 0.698      & 0.534     & 0.740     & 0.656     & 0.514     & 0.744      & 0.689      & 0.519     & 0.715       & 0.640      & 0.486       \\
Self-Expression   & 0.728      & 0.673      & 0.512     & 0.775      & 0.682      & 0.536     & 0.773     & 0.687     & 0.556     & 0.779      & 0.720      & 0.585     & 0.740       & 0.667      & 0.525       \\
Interpersonal     & 0.499      & 0.510      & 0.309     & 0.503      & 0.532      & 0.318     & 0.486     & 0.525     & 0.307     & 0.428      & 0.486      & 0.269     & 0.450       & 0.503      & 0.265       \\
Decision Making   & 0.477      & 0.472      & 0.260     & 0.433      & 0.448      & 0.238     & 0.475     & 0.492     & 0.294     & 0.455      & 0.497      & 0.305     & 0.468       & 0.483      & 0.267       \\
Stress Management & 0.375      & 0.444      & 0.210     & 0.283      & 0.362      & 0.141     & 0.342     & 0.444     & 0.196     & 0.305      & 0.375      & 0.161     & 0.336       & 0.371      & 0.170       \\
Deficit           & 0.517      & 0.532      & 0.325     & 0.395      & 0.492      & 0.241     & 0.481     & 0.543     & 0.318     & 0.426      & 0.508      & 0.278     & 0.446       & 0.481      & 0.283       \\ \bottomrule
\end{tabular}
\end{center}
\caption{Evaluation results under different persona conditions
for Qwen3-Omni-30B. Within each EQ composite area, the three
columns report $\mathrm{Acc}_1$, $\mathrm{Acc}_2$, and
$\mathrm{Acc}_{\mathrm{traj}}$, respectively, from left to right.}\label{tab:eq_scale}
\end{table*}

\end{document}